\documentclass[11pt]{article}

\usepackage[final]{acl}

\usepackage{times}
\usepackage{latexsym}

\usepackage[T1]{fontenc}

\usepackage[utf8]{inputenc}

\usepackage{microtype}

\usepackage{inconsolata}

\usepackage{graphicx}
\usepackage{hyperref}
\usepackage{xcolor}
\usepackage{multirow}
\usepackage{multicol}
\usepackage{makecell}
\usepackage{wrapfig}
\usepackage{utfsym}
\usepackage{bm}
\usepackage{amsmath}
\usepackage{amssymb}
\usepackage{mathtools}
\usepackage{amsthm}
\usepackage{xspace}
\usepackage{soul}
\usepackage{enumerate}
\usepackage{subcaption}
\usepackage{booktabs}
\usepackage{pdfrender}
\usepackage{tcolorbox}
\tcbuselibrary{most}

\newcommand{\ourapproach}{\textsc{Func-R1}\xspace}
\newcommand{\geoapproach}{\textsc{Geo-R1}\xspace}
\newcommand{\sciapproach}{\textsc{SynSci-R1}\xspace}
\newcommand{\captiondataset}[0]{\texttt{FuncCaption}}
\newcommand{\reasoningdataset}[0]{\texttt{FuncReasoning}}
\newcommand{\Caption}{\captiondataset\xspace}
\newcommand{\Reasoning}{\reasoningdataset\xspace}

\definecolor{myred}{HTML}{FFB3B3}
\definecolor{mypurple}{HTML}{AEB7FF}
\definecolor{myyellow}{HTML}{FFE5B7}
\definecolor{mygreen}{HTML}{9EFAB5}

\definecolor{ForestGreen}{RGB}{0,235,110}

\newcommand{\hlred}[1]{\sethlcolor{myred}\hl{#1}}
\newcommand{\hlpurple}[1]{\sethlcolor{mypurple}\hl{#1}}
\newcommand{\hlyellow}[1]{\sethlcolor{myyellow}\hl{#1}}
\newcommand{\hlgreen}[1]{\sethlcolor{mygreen}\hl{#1}}

\definecolor{backred}{RGB}{255, 190, 190}
\definecolor{backblue}{RGB}{220, 230, 250}

\newtcolorbox{promptbox}[1]{%
  enhanced,
  breakable,
  colback=gray!5,
  colframe=gray!50,
  colbacktitle=gray!55,
  coltitle=white,
  fonttitle=\bfseries\small,
  title={#1},
  left=6pt, right=6pt, top=4pt, bottom=4pt,
  boxrule=0.5pt,
}

\tcbset{
  apiCard/.style={
    colback=white,
    colframe=blue!60!black,    
    colbacktitle=blue!10!white, 
    coltitle=black,
    boxrule=0.8pt,
    arc=1mm,
    width=\textwidth,
    left=6pt, right=6pt, top=1pt, bottom=1pt,
    fonttitle=\bfseries\sffamily\footnotesize,
    title=#1,
    after skip=0.02cm
  }
}

\tcbset{
  openCard/.style={
    colback=white,
    colframe=green!60!black,   
    colbacktitle=green!10!white,
    coltitle=black,
    boxrule=0.8pt,
    arc=1mm,
    width=\textwidth,
    left=6pt, right=6pt, top=1pt, bottom=1pt,
    fonttitle=\bfseries\sffamily\footnotesize,
    title=#1,
    after skip=0.02cm
  }
}

\title{\ourapproach 
  \raisebox{-0.4cm}{\includegraphics[width=1.2cm, height=1.2cm]{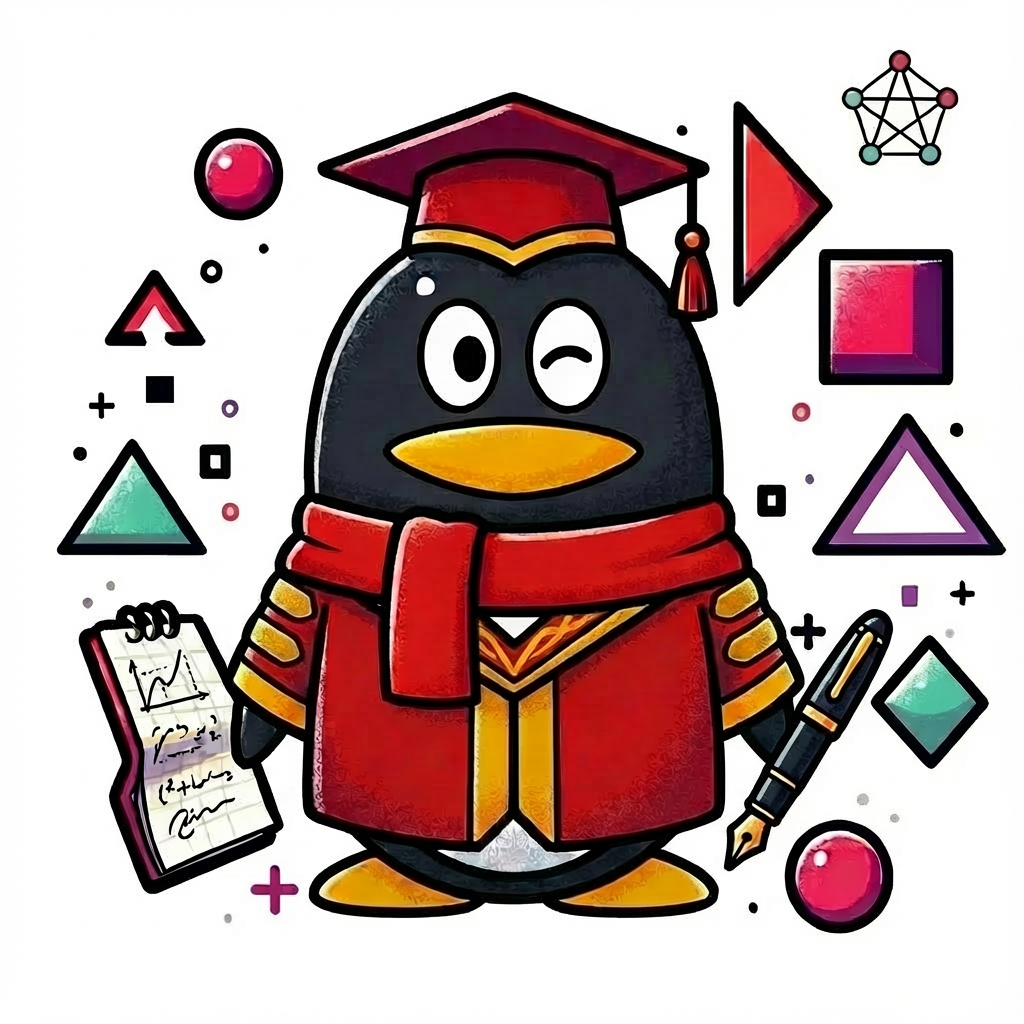}}: 
  Incentivizing Mathematical Function Reasoning \\ in Multimodal Large Language Models
  \vspace{0.5em}}

\author{
 \textbf{Mingze Yin\textsuperscript{1,2,*}},
 \textbf{Xiaohan Wang\textsuperscript{2,*}},
 \textbf{Dian Li\textsuperscript{2,$\dagger$}},
 \textbf{Haichao Yao\textsuperscript{1,2}},
 \textbf{Yilin Zhao\textsuperscript{2}},
 \textbf{Youjun Chen\textsuperscript{2}},
\\
 \textbf{Gang Liu\textsuperscript{2}},
 \textbf{Jintai Chen\textsuperscript{3}},
 \textbf{Yiheng Zhu\textsuperscript{4,$\dagger$}},
 \textbf{Chang-Yu Hsieh\textsuperscript{5,$\dagger$}}
 \textbf{Aimin Pan\textsuperscript{6,7,$\dagger$}}
 \vspace{1em}
\\
 \textsuperscript{1}College of Computer Science and Technology, Zhejiang University \\
 \textsuperscript{2}Tencent 
 \textsuperscript{3}AI Thrust, HKUST (GZ)
 \textsuperscript{4}Zhongguancun Academy \\
 \textsuperscript{5}College of Pharmaceutical Sciences, Zhejiang University 
 \textsuperscript{6}VNET Group
 \textsuperscript{7}Zhejiang Lab
 \vspace{0.3em}
\\
 \small{\textsuperscript{*}Equal Contribution.\quad
         \textsuperscript{$\dagger$}Corresponding Authors.}
}

\begin{document}
\maketitle

\begingroup
  \renewcommand\thefootnote{}%
  \footnotetext{\textbf{Correspondence to:}
    \href{mailto:mzyin256@gmail.com}{mzyin256@gmail.com}, 
    \href{mailto:kimhsieh@zju.edu.cn}{kimhsieh@zju.edu.cn},
    \enspace\href{mailto:pan.aimin@vnet.com}{pan.aimin@vnet.com}}%
  \addtocounter{footnote}{-1}%
\endgroup

\begin{abstract}
Performing deliberate mathematical reasoning in visual contexts is a hallmark of advanced Multimodal Large Language Models (MLLMs) and requires a sophisticated synthesis of perceptual grounding and symbolic logic. 
However, in the realm of mathematical functions, our investigation reveals a critical \textit{modality interference phenomenon}: even advanced models, while performing textual computational reasoning, tend to disregard or misinterpret essential visual cues.
To address this challenge, we propose \textbf{\ourapproach}, which synergistically harmonizes precise visual perception and rigorous logical reasoning.
Concretely, built upon an explicitly decoupled architecture, we employ a hierarchical post-training framework to progressively identify critical visual evidence and conduct in-depth theoretical reasoning.
Furthermore, the \textit{Perception-Aligned Theoretic Optimization (PATO)} strategy is proposed to steer policy updating towards internalizing fundamental theoretical properties while dynamically rectifying heterogeneous visual information throughout the reasoning process.
Extensive experiments across diverse benchmarks demonstrate that \ourapproach delivers the optimal performance among open-source MLLMs, even surpassing GPT-5 with an 8.4\% improvement on MathVerse's function-oriented tasks.
\end{abstract}

\section{Introduction}
Mathematical reasoning in visual contexts represents the sophistication of human intelligence, as it hinges on precise visual interpretation, substantive domain knowledge, and rigorous logical inference. Significant advances in Multimodal Large Language Models (MLLMs) have spurred discourse on the prospect of demonstrating precise mathematical reasoning, advancing towards Artificial General Intelligence (AGI)~\citep{AlphaGeometry, groupon, Caduceus, MathCoder-VL}. 

Mathematical functions, grounded in abstraction-centric symbolic formalisms, do not abide by the same constraints that govern conventional problems in geometry, statistics, topology or combinatorics.  
The function-driven computational deduction demands global coherence and quantitative exactitude, whereas diagram parsing emphasizes localized perception derived from limited and approximate information.
Pronounced information asymmetries compel the model to encode bifurcating trajectories, thereby hindering the reconciliation of function-oriented diagram interpretation and computational reasoning~\citep{math_theory1}.

\begin{figure}[!ht]
    \centering
    \includegraphics[width=0.80\linewidth]{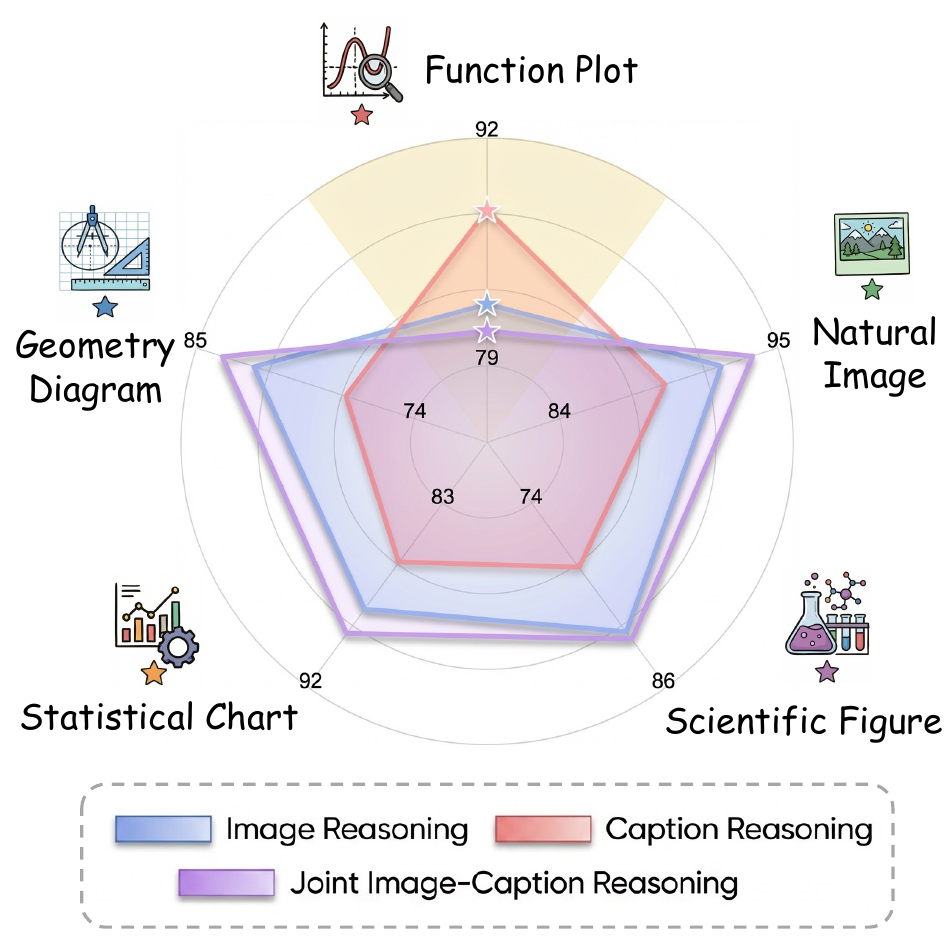}
    \caption{GPT-5’s mathematical reasoning performance trends under diverse multimodal conditions vary systematically across distinct visual contexts.}
    \label{figure1}
    \vspace{-0.5em}
\end{figure}

Herein, a simple and intuitive heuristic experiment is conducted to reveal the domain-specific \textit{modality interference phenomenon}. 
Specifically, we employ the most powerful LLM to date GPT-5 to execute reasoning over diverse problem types from MathVista~\citep{MathVista}. 
Three inference settings are specified: 
\emph{(i)} supplying the diagram (Image Reasoning), 
\emph{(ii)} supplying a textual description of the diagram (Caption Reasoning), 
and \emph{(iii)} supplying both (Joint Image-Caption Reasoning). 
As illustrated in Figure~\ref{figure1}, while performance trends match expectations for most problem types, reasoning with function diagrams is notably inferior to reasoning with textual descriptions. 
Paradoxically, jointly providing synergistic visual diagrams and textual descriptions yields the worst performance.
This anomalous pattern indicates that MLLMs are prone to selectively disregard or partially forget crucial mathematical visual clues when engaging in text-based computational reasoning. Such intrinsic irreconcilability constitutes a major impediment to incentivizing deliberate function reasoning.

\begin{figure*}[!t]
    \centering
    \includegraphics[width=0.98\textwidth]{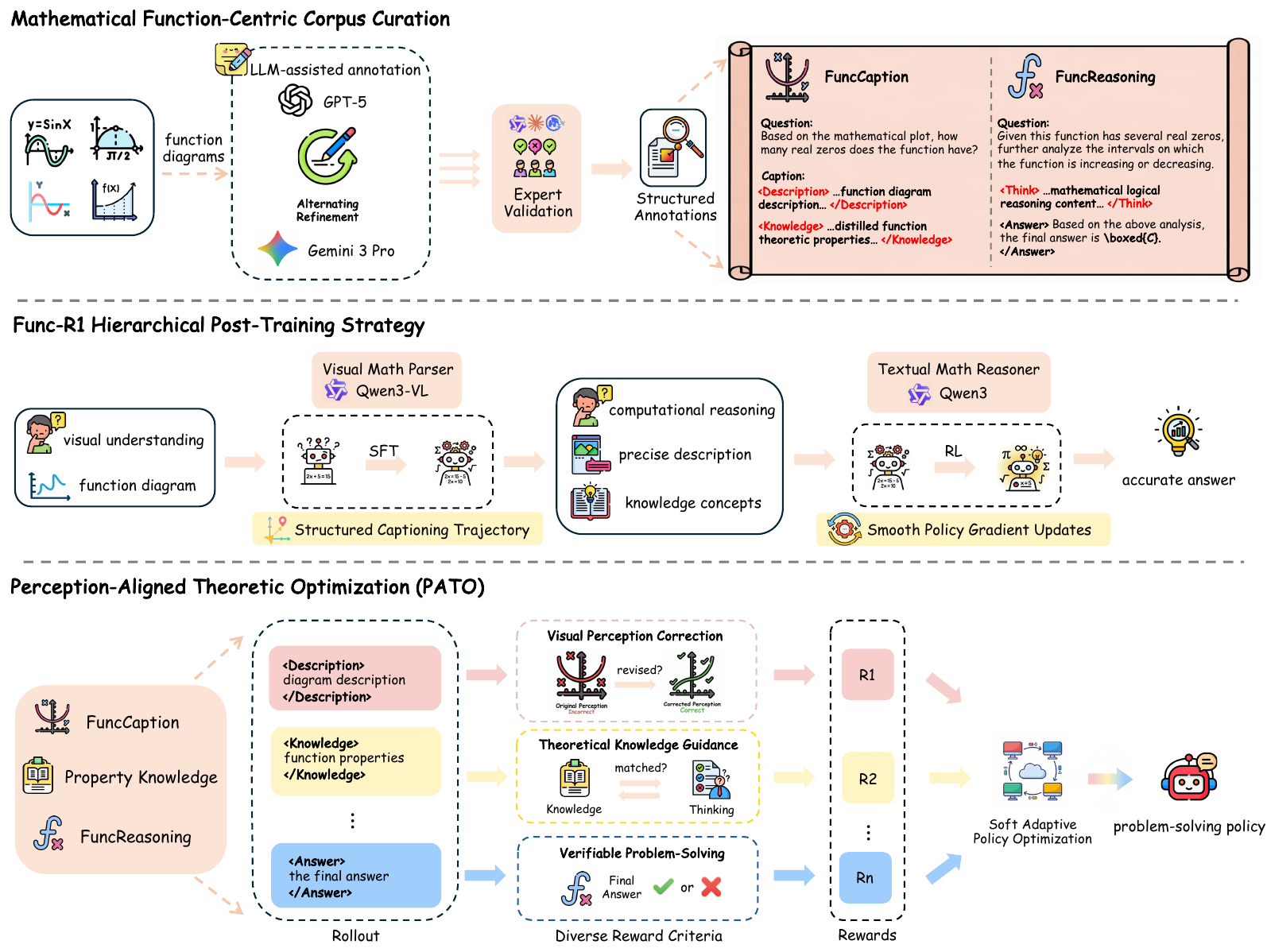}
    \caption{\textbf{Overview of the \ourapproach framework}. 
    We adopt a progressive optimization approach that combines SFT and RL under an explicitly decoupled architecture to integrate visual perception and logical reasoning capabilities.}
    \label{figure2}
\end{figure*}

To overcome such interference issue, an explicitly decoupled model architecture is designed, consisting of a visual math parser and a textual math reasoner.
We curate function-oriented mathematical corpora comprising 3.5k problems, with annotations capturing both visual understanding and logical reasoning, constructed through an LLM-enhanced automated pipeline and validated by human mathematics experts.
In the hierarchical training framework, we first perform structured Supervised Fine-Tuning (SFT) for the visual math parser. 
The systematic captioning trajectory encourages the model to move beyond basic visual element extraction and capture function properties, fostering deeper visual understanding.
In the subsequent training stage, we propose the Perception-Aligned Theoretic Optimization (PATO) strategy to systematically enhance the rigor and depth of a textual math reasoner’s computational deduction.
In contrast to conventional Group Relative Policy Optimization (GRPO), we adopt a smooth and adaptive policy optimization scheme to balance exploitation and exploration, yielding stable and performant Reinforcement Learning (RL) policy updating.
Furthermore, the customized reward signals progressively incentivize the model to rectify visual perception anomalies, assimilate sophisticated function-theoretic properties, and execute nuanced computational reasoning.
Eventually, the collaboration of visual math parser and textual math reasoner involves revisiting visual evidence and leveraging theoretical knowledge throughout the problem-solving process, therefore catalyzing advanced reasoning over mathematical functions.



Our contributions can be summarized as follows:
\begin{itemize}
\vspace{-0.3em}
\item[$\bullet$]
\textit{Modality Interference Issue.} 
We reveal intrinsic modality interference in the realm of mathematical functions and leverage an explicitly decoupled architecture and a hierarchical post-training strategy to alleviate it.
\vspace{-0.5em}
\item[$\bullet$]
\textit{Curated Diverse Corpora.} 
We construct 3.5k function-centric corpora with LLM-assisted annotation and human validation, supporting the optimization of perception and reasoning.

\vspace{-0.5em}
\item[$\bullet$]
\textit{Innovative Policy Optimization.} The tailored PATO is a training-stable, online-decidable policy update mechanism with hybrid rewards, coupling visual interpretation calibration, theoretical knowledge guidance, and verifiable computational reasoning.
\vspace{-0.5em}
\item[$\bullet$]
\textit{Substantial Reasoning Enhancements.} \ourapproach attains open-source state-of-the-art reasoning performance on five benchmarks, notably outperforming GPT-5 by 8.4\% on MathVerse.
\end{itemize}

\section{Method}
As illustrated in Figure~\ref{figure2}, an explicitly decoupled architecture is incorporated to facilitate the integration of visual perception and logical reasoning. 
We enhance the visual perception capability of the visual math parser through SFT (Sec.~\ref{SFT}), and improve the logical reasoning ability of the textual math reasoner via RL (Sec.~\ref{RL}).

\subsection{Mathematical Corpus Construction}
\label{Corpus}
We collect over 10k mathematical function problems from publicly authoritative mathematics websites. The MinerU framework~\citep{MinerU} is first employed for data preprocessing, including problem parsing and function plot extraction. 
To enrich the annotations, we leverage GPT-5 and Gemini-3-Pro to generate image captions, identify relevant knowledge concepts, and refine reasoning trajectories. 
In addition, Qwen3-VL, Claude Opus 4.1, and GLM-4.1V are adopted as arbitration models to verify the annotations and resolve discrepancies.
Finally, five Ph.D. students majoring in mathematics are recruited to perform manual validation, focusing on the visual evidence, key formulas, and logical soundness of the reasoning process.
Eventually, two function-centric datasets comprising 3,519 high-quality, well-annotated mathematical problems are constructed: \Caption, comprising high-quality diagram captions, and \Reasoning, consisting of function-oriented question–answer pairs.

\subsection{Domain Knowledge Enhancement}
\label{SFT}
We begin with the optimization of visual math parser to precisely perceive and interpret mathematical function diagrams through domain-specific Supervised Fine-Tuning (SFT). 
Based on the problem statement and function diagram, the structured captioning trajectory is organized as: 
\texttt{<Description>...</Description> <Property>...</Property>.}
By elaborating progressive reasoning captions, we empower the MLLM to faithfully comprehend intricate mathematical properties pertinent to subsequent problem-solving.
Leveraging the curated \Caption for SFT, we employ a standard cross-entropy loss to endow visual math parser with the structural captioning format and function-theoretic knowledge. 
This SFT phase provides a structured cognitive prior, improving the visual perception and knowledge comprehension for downstream mathematical problem-solving tasks.

\subsection{Perception-Aligned Theoretic Optimization (PATO)}
\label{RL}
\subsubsection{Smooth Policy Gradient Updates}
Under conventional Group Relative Policy Optimization (GRPO)~\citep{GRPO}, hard gradient clipping induces high-variance advantage estimates, which in turn constrain the attainable convergence performance.
As a refinement, Soft Adaptive Policy Optimization (SAPO)~\citep{SAPO} adaptively attenuates off-policy updates to maintain a sustained and stable optimization signal, but often at the expense of slower convergence.
Inspired by SAPO, our PATO adapts a smooth, temperature-controlled gating mechanism to rich and informative reward learning signals, enabling stable and efficient RL post-training.
(See Sec.~\ref{ablation_study}, Policy Gradient Updates, for convergence comparisons.)
\vspace{-0.5em}
\begin{multline}
    \mathcal{L}_{\operatorname{RL}}(\theta)
    = \mathbb{E}_{q\sim\mathcal{D},\,\{y_i\}_{i=1}^{G}\sim \pi_{\theta_{\mathrm{old}}}(\cdot\mid q)} \\
    \left[\frac{1}{G}\sum_{i=1}^{G}\frac{1}{|y_i|}\sum_{t=1}^{|y_i|} f_{i,t}\!\left(r_{i,t}(\theta)\right)\,\hat{A}_{i,t}\right],
\end{multline}
\begin{equation}
    \widehat{A}_{i,t} = \widehat{A}_i = \frac{R_i - \mathtt{mean}(\{R_j\}_{j=1}^G)}{\mathtt{std}\left(\{R_j\}_{j=1}^G\right)},
\end{equation}

Concretely, a token-level soft trust region is introduced to enhance sequence-level coherence, and an asymmetric temperature scheme is adopted to accommodate the differing dynamics of positive versus negative token updates.
\begin{equation}
    f_{i,t}(x)=\sigma\!\left(\tau_{i,t}(x-1)\right)\cdot\frac{4}{\tau_{i,t}},
\end{equation}
\begin{equation}
    \tau_{i,t} = 
    \begin{cases} 
    \tau_{\mathrm{pos}}, & \text{if}\hat{A}_{i,t}>0, \\
    \tau_{\mathrm{neg}}, & \text{otherwise}. 
    \end{cases}
\end{equation}
where $\tau_{\mathrm{pos}}$ and $\tau_{\mathrm{neg}}$ are the temperatures for positive and negative tokens, respectively.

\subsubsection{Reward Signal Design}
\label{reward_signal_design}
To effectively guide the model’s mathematical reasoning optimization, we design a multi-component reward that integrates \textit{robust visual parsing, profound knowledge understanding, and rigorous computational reasoning}.
The proposed mathematical problem-solving policy optimization framework comprehensively incorporates rule-based, LLM-assisted, and directly verifiable rewards. 

\textbf{Visual Perception Correction Reward.}
We incentivize the problem-solving model to rectify heterogeneous visual perceptual information, given the intrinsic complexity and observational ambiguities of function plots.
LLM-as-a-Judge framework is incorporated to determine whether the model successfully avoids, detects, or corrects visual perception errors during the mathematical reasoning process.
Concretely, Qwen3-VL-30B-A3B~\citep{Qwen3} is invoked to produce discrete labels rather than a free-form scalar score. 
\textit{(i)} $r \in \{0,1\}$ indicates model directly refers to the original caption; 
\textit{(ii)} $d \in \{0,1\}$ indicates model identifies caption errors and selectively ignores parts; 
\textit{(iii)} $c \in \{0,1\}$ indicates model precisely detects caption errors and explicitly corrects them.
The final reward is deterministically assigned through a fixed rule-based mapping. 
\begin{equation}
    \mathcal{R}_{v} = 
    \begin{cases}
    2, & c=1,\\
    1, & c=0,\ d=1,\\
    -1, & d=0,\ r=1,\\
    0, & \text{otherwise}.
    \end{cases}
\end{equation}
This design reduces sensitivity to prompt variations, enhances reproducibility, and provides stable reward signals that incentivize the model to proactively engage in visual perception correction.


\textbf{Theoretical Knowledge Guidance Reward.}
We further encourage the model to generate rigorous reasoning processes guided by underlying mathematical principles.
Let $K_q$ and $K_r$ denote the knowledge items associated with the question and invoked in the reasoning trajectory, respectively. 
Intuitively, high-quality reasoning should cover the key theoretical concepts needed for solving current problems, while avoiding irrelevant or spurious knowledge. We therefore measure global knowledge alignment with the F1 score $\mathrm{F}_1(K_q, K_r)$, which captures both the coverage and precision of the theoretical knowledge.
Additionally, to prevent the model from merely mentioning relevant concepts without effectively applying them in reasoning, we incorporate the maximum similarity between intermediate reasoning steps and theoretical knowledge items to characterize semantic alignment at a finer granularity.
\begin{equation}
    \mathcal{R}_k= \mathrm{F}_1(K_q, K_r) \cdot \mathbb{E}_{k \sim K_q \cap K_r} \left[ \mathtt{sim}(k_q, k_r) \right].
\end{equation}
This reward formulation combines global knowledge matching with local knowledge utilization, jointly promoting overall consistency and logically productive knowledge application throughout intricate reasoning.

\textbf{Verifiable Problem-Solving Reward.}
Building on well-calibrated visual parsing information and thorough theoretical knowledge comprehension, we incentivize the model to engage in structured inference and profound computational reasoning to achieve correct solutions.
The solving model is required to generate responses that adhere to a strict format containing a reasoning segment and an answer segment (\textit{e.g.}, \mbox{\texttt{<Think>...</Think> <Answer>...</Answer>}}). 
Moreover, the reward for problem-solving correctness is standardized owing to the inherently verifiable nature of mathematics~\citep{Tulu3}.
\begin{equation}
    \mathcal{R}_{\operatorname{s}} =
    \begin{cases}
    \displaystyle \mathbf{1}\!\left(y_i = \hat{y_i}\right), 
    & \text{valid format},\\
    -1, & \text{invalid format}.
    \end{cases}
\end{equation}
where $\hat{y}_i \in \hat{\mathbf{y}} = [\hat{y}_1, \ldots, \hat{y}_n]$ represents the extracted predicted answer, $y_i \in \mathbf{y} = [y_1, \ldots, y_n]$ indicates the reference solution, and $\mathbf{1}(\cdot)$ denotes the indicator function. 
The textual math reasoner is incentivized to execute rigorous computational derivation grounded in structured reasoning, facilitating mathematical problem-solving efficacy.

\begin{table*}[!htb]
\centering
\resizebox{0.98\linewidth}{!}{
    \begin{tabular}{lcccc c cccc@{\hspace{0.3em}}c}
    \toprule
    \multirow{2}{*}{\textbf{Model}} & \multicolumn{4}{c}{\textbf{MathVista}} & \multicolumn{1}{c}{\textbf{Math-V}} & \multicolumn{4}{c}{\textbf{MathVerse}} & \multicolumn{1}{c}{\textbf{We-Math}} \\
    \cmidrule(lr){2-5} \cmidrule(lr){6-6} \cmidrule(lr){7-10} \cmidrule(lr){11-11}
    & \textbf{Disc.} & \textbf{Lin.} & \textbf{Nonlin.} & \textbf{Comp.} & \textbf{Func.} & \textbf{Disc.} & \textbf{Lin.} & \textbf{Nonlin.} & \textbf{Comp.} & \textbf{Func.} \\
    \midrule
    \multicolumn{11}{c}{\textit{Closed-Source Multimodal Large Language Models (MLLMs)}} \\
    \midrule
    GPT-4V 
    & 52.3 & 50.5 & 51.8 & 47.2 
    & 36.1
    & 41.2 & 38.0 & 35.1 & 33.6 
    & 45.6 \\
    GPT-4o 
    & 60.3 & 59.2 & 64.7 & 60.3 
    & 37.0 
    & 40.5 & 39.9 & 37.0 & 35.6 
    & 50.1 \\
    Qwen-VL-Max 
    & 63.2 & 62.6 & 60.6 & 54.8 
    & 40.7 
    & 29.1 & 27.8 & 26.8 & 22.8 
    & 49.8 \\
    Gemini 2.5 Flash 
    & 83.2 & 80.4 & 77.2 & 73.4 
    & 53.2
    & 58.2 & 56.8 & 51.0 & 49.8 
    & 56.6 \\
    Gemini 3 Flash 
    & 84.1 & 80.0 & 78.5 & 75.3 
    & 51.8
    & 60.2 & 58.3 & 54.4 & 52.1 
    & 59.7 \\
    GPT-5 
    & \colorbox{gray!25}{84.7} 
    & \colorbox{gray!25}{81.0} 
    & \colorbox{gray!25}{\textbf{79.9}} 
    & \colorbox{gray!25}{76.1} 
    & \colorbox{gray!25}{\textbf{57.0}} 
    & \colorbox{gray!25}{61.6} 
    & \colorbox{gray!25}{60.4} 
    & \colorbox{gray!25}{56.3} 
    & \colorbox{gray!25}{55.6} 
    & \colorbox{gray!25}{67.3} \\
    \midrule
    \multicolumn{11}{c}{\textit{Open-Source Multimodal Large Language Models (MLLMs)}} \\
    \midrule
    LLaVA-V1.5-7B 
    & 22.7 & 23.6 & 10.2 & 27.2 
    & 14.7
    & 17.6 & 10.4 & 11.6 & 12.6 
    & 27.9 \\
    LLaVA-V1.5-13B 
    & 27.7 & 25.6 & 22.1 & 21.3 
    & 18.0 
    & 18.5 & 14.2 & 15.9 & 13.5 
    & 23.6 \\
    InternVL2-8B 
    & 34.2 & 34.0 & 35.4 & 32.5 
    & 11.3
    & 23.1 & 21.7 & 22.7 & 18.2 
    & 33.0 \\
    InternVL2-40B 
    & 38.4 & 37.2 & 33.4 & 30.7 
    & 19.5 
    & 25.1 & 24.0 & 21.7 & 19.2 
    & 38.5 \\
    DeepSeek-VL2 
    & 60.0 & 57.0 & 55.7 & 53.8 
    & 30.8 
    & 43.8 & 42.1 & 43.0 & 37.5 
    & 39.3 \\
    GLM-4.1V 
    & 63.3 & 62.7 & 65.2 & 68.3 
    & 36.2 
    & 44.2 & 40.3 & 43.7 & 39.9 
    & 40.4 \\
    Qwen3-VL-8B 
    & 66.8 & 65.3 & 60.1 & 62.7 
    & 39.3 
    & 47.8 & 45.3 & 44.1 & 42.7 
    & 47.3 \\
    Qwen3-VL-32B 
    & 70.3 & 68.4 & 65.9 & 62.1 
    & 42.1 
    & 53.3 & 51.7 & 50.5 & 44.8 
    & 58.1 \\
    Math-LLaVA 
    & 49.8 & 47.6 & 43.7 & 42.4 
    & 23.9
    & 23.6 & 28.4 & 19.7 & 19.3  
    & 37.7 \\
    R1-VL 
    & 69.6 & 66.0 & 65.4 & 66.4 
    & 36.7 
    & 54.3 & 50.5 & 50.3 & 45.9 
    & 55.7 \\
    MathCoder2
    & 73.9 & 70.4 & 67.8 & 66.9
    & 45.8
    & 56.0 & 54.7 & 50.9 & 50.1
    & 60.1 \\
    Vision-R1
    & 75.7 & 72.6 & 69.9 & 68.8
    & 46.9
    & 56.8 & 55.9 & 51.8 & 51.0
    & 61.1 \\
    CodePercept
    & 77.2 & \colorbox{backblue!75}{74.1} & 71.3 & 70.2
    & \colorbox{backblue!75}{48.2}
    & 58.3 & 57.4 & 53.5 & 52.7
    & 62.6 \\
    \midrule
    \multicolumn{11}{c}{\textbf{\ourapproach-Lite:} \textit{Qwen3-VL-4B SFT + Qwen3-8B SAPO}} \\
    \midrule
    SFT 
    & 71.5 
    & 69.7 
    & 69.0 
    & 63.3 
    & 35.7 
    & 52.8 
    & 47.0 
    & 51.2 
    & 43.5 
    & 57.1 \\
    SFT + PATO    
    & \colorbox{backblue!75}{77.4} 
    & 73.6 
    & \colorbox{backblue!75}{73.3} 
    & \colorbox{backblue!75}{70.6} 
    & 42.4 
    & 58.9 
    & 58.7 
    & \colorbox{backblue!75}{57.0} 
    & 55.3 
    & \colorbox{backblue!75}{64.8} \\
    \midrule
    \multicolumn{11}{c}{\textbf{\ourapproach-Pro:} \textit{Qwen3-VL-8B SFT + Qwen3-32B SAPO}} \\
    \midrule
    SFT 
    & 76.2 
    & \colorbox{backblue!75}{74.1} 
    & 73.0 
    & 66.5 
    & 46.9 
    & \colorbox{backblue!75}{60.2} 
    & \colorbox{backblue!75}{60.5} 
    & 56.8 
    & \colorbox{backblue!75}{57.7} 
    & 61.2 \\
    SFT + PATO
    & \colorbox{backred!50}{\textbf{85.3}}
    & \colorbox{backred!50}{\textbf{82.6}} 
    & \colorbox{backred!50}{79.3} 
    & \colorbox{backred!50}{\textbf{78.0}} 
    & \colorbox{backred!50}{56.4} 
    & \colorbox{backred!50}{\textbf{66.1}} 
    & \colorbox{backred!50}{\textbf{65.7}} 
    & \colorbox{backred!50}{\textbf{61.2}}
    & \colorbox{backred!50}{\textbf{60.5}}
    & \colorbox{backred!50}{\textbf{70.8}} \\
    \bottomrule
    \end{tabular}}
    \caption{\textbf{Out-of-distribution function reasoning performance over diverse mathematical benchmarks.} 
    The optimal and suboptimal open-source models are highlighted in \colorbox{backred!50}{red} and \colorbox{backblue!75}{blue}, respectively.
    We highlight the best results, including closed-source models, in \textbf{bold}.}
    \label{table3}
\end{table*}

\subsubsection{Overall Training Objective}
During the RL phase, we aim to optimize the policy model $\pi_\theta$ utilizing the SAPO training paradigm. The crux of this procedure is the design of the reasoning reward signal, which directs the model’s learning process. 
The overall reward $\mathcal{R}_{\operatorname{PATO}}$, used to compute the group-wise relative advantages, is formulated as a weighted combination of the multifaceted components detailed in Sec.~\ref{reward_signal_design}.
\begin{equation}
    \mathcal{R}_{\operatorname{PATO}} = 
    \lambda_{1} \mathcal{R}_{\operatorname{v}} + 
    \lambda_{2} \mathcal{R}_{\operatorname{k}} + \mathcal{R}_{\operatorname{s}}.
\end{equation}
These non-negative weight coefficients $\lambda_{1}$ and $\lambda_{2}$ serve as critical hyperparameters for calibrating the model, allowing us to modulate the emphasis among dynamically refining visual analysis, promoting understanding of theoretical knowledge, and preserving a coherent reasoning process.

\section{Experiments}
In the experimental section, we progressively present the following findings:
\textit{(i)} \ourapproach achieves the best function problem-solving results on the internal test set, with an average accuracy of 91.4\% (Sec.~\ref{in-distribution}).
\textit{(ii)} \ourapproach attains outstanding reasoning performance on five mainstream mathematical benchmarks comparable to that of the powerful closed-source GPT-5 (Sec.~\ref{out-of-distribution}).
\textit{(iii)} The training approach is further extended to other reasoning domains, such as geometry, to demonstrate strong universality (Sec.~\ref{generalized-reasoning}).

\subsection{Experimental Setups}
We build two model variants, \ourapproach-Lite and \ourapproach-Pro, based on the Qwen3 model family.
Using a hierarchical post-training strategy, we progressively optimize the visual math parser and the textual math reasoner. Specifically, \Caption is used entirely for SFT, while \Reasoning is split into 75\% / 10\% / 15\% partitions to support RL optimization and in-distribution evaluation.
Training details are provided in Appendix~\ref{implement_details}, while the LLM baselines and evaluation benchmarks are described in Appendix~\ref{experiment_details}.
The code and data will be publicly available at \url{https://github.com/zju-ai4s/Func-R1} upon completion of the company's disclosure review process.

\subsection{In-Distribution Function Reasoning}
\label{in-distribution}
To rigorously interrogate the mathematical problem‑solving competence of MLLMs involving visual diagrams, we begin with function reasoning assessments on the in-distribution \Reasoning test set. 
Evaluation data is meticulously classified into four function types: Discrete Function (Disc. Func.), Linear Function (Lin. Func.), Nonlinear Function (Nonlin. Func.), and Composite Function (Comp. Func.). As presented in Table~\ref{table2}, with increasing function complexity, all models exhibit a declining trend in problem‑solving efficacy.
\ourapproach attains state-of-the-art in-distribution function reasoning, with an average accuracy of 91.4, surpassing closed-source LLMs, including Gemini 3 Flash, Claude Opus 4.7, and GPT-5.

\begin{table}[htbp]
    \centering
    \resizebox{\columnwidth}{!}{
    \begin{tabular}{l|c@{\hspace{1em}}c@{\hspace{0.8em}}c@{\hspace{0.5em}}c}
        \toprule
        \textbf{Model} & \textbf{Disc. Func.} & \textbf{Lin. Func.} & \textbf{Nonlin. Func.} & \textbf{Comp. Func.} \\
        \midrule
        \multicolumn{5}{c}{\textit{Closed-Source MLLMs}} \\
        \midrule
        GPT-4o & 77.0 & 76.4 & 72.9 & 69.5 \\
        Gemini 2.5 Flash & 80.9 & 82.2 & 79.4 & 75.9 \\
        Gemini 3 Flash & 84.8 & 84.6 & 83.3 & 78.1 \\
        Claude Opus 4.1 & 82.1 & 77.5 & 76.4 & 73.8 \\ 
        Claude Opus 4.7 
        & 89.3 
        & \colorbox{gray!25}{91.3} 
        & \colorbox{gray!25}{88.2} 
        & 79.9 \\ 
        GPT-5 
        & \colorbox{gray!25}{91.5}
        & {89.0}
        & 82.6
        & \colorbox{gray!25}{83.4} \\
        \midrule
        \multicolumn{5}{c}{\textit{Open-Source MLLMs}} \\
        \midrule
        CogVLM2 & 74.4 & 70.2 & 65.4 & 58.7 \\
        WeThink & 75.0 & 71.8 & 62.0 & 59.2 \\
        Math-LLaVA & 74.8 & 70.5 & 67.1 & 64.9 \\
        GLM4.1-V & 79.3 & 72.7 & 71.4 & 68.8 \\
        Qwen3-VL-8B & 78.3 & 74.8 & 70.5 & 67.0\\
        Qwen3-VL-32B & 82.6 & 81.1 & 79.7 & 75.4 \\
        \midrule
        \textbf{\ourapproach-Lite} 
        & \colorbox{backblue!75}{92.7} 
        & \colorbox{backblue!75}{89.4} 
        & \colorbox{backblue!75}{84.5} 
        & \colorbox{backblue!75}{82.9} \\
        \textbf{\ourapproach-Pro} 
        & \colorbox{backred!50}{\textbf{94.3}} 
        & \colorbox{backred!50}{\textbf{92.7}} 
        & \colorbox{backred!50}{\textbf{90.0}} 
        & \colorbox{backred!50}{\textbf{88.5}} \\
        \bottomrule
    \end{tabular}}
    \caption{
    \textbf{In-distribution function reasoning performance} covering granular function classes.}
    \label{table2}
\end{table}

\subsection{Out-Of-Distribution Function Reasoning}
\label{out-of-distribution}
To further validate the proposed paradigm confers broad improvements in mathematical function problem-solving, we holistically conduct more challenging out-of-distribution function reasoning evaluations. As presented in Table~\ref{table3}, \ourapproach significantly advances the state-of-the-art performance on 8 of 10 evaluation suites. Notably, \ourapproach even outperforms GPT-5 across all MathVerse function subcategory reasoning tasks, yielding an average relative improvement of 8.4\%. The exceptional performance across numerous mathematical reasoning benchmarks further validates the effectiveness of the proposed training strategy and optimization algorithm.

\subsection{Generalized Multi-Discipline Reasoning}
\label{generalized-reasoning}
We further explore whether the proposed training framework could enhance MLLMs' reasoning capabilities across diverse disciplinary contexts.
Under the proposed training framework, \geoapproach is trained on GeoReasoning-10K, whereas \sciapproach is trained on \Caption, \Reasoning and GeoReasoning-10K, adhering to the identical base models (\textit{i.e.}, Gemma3-4B and Qwen2.5-7B-Instruct) as in~\citet{GeoReasoning}.
The generalized reasoning performance is thoroughly assessed across seven college-level core disciplines with the MathVista~\citep{MathVista} and MMMU~\citep{MMMU} databases.
Extensive commonly-used geometric baselines are incorporated, including Gemma3~\citep{Gemma3}, AutoGeo~\citep{AutoGeo}, GeoPeP~\cite{MathGlance}, GeoGPT4V~\citep{GeoGPT4V}, Geo170K~\citep{G-LLaVA}, and GeoReasoning~\cite{GeoReasoning} to facilitate comprehensive comparison. 
Table~\ref{table4} shows that \geoapproach delivers outstanding geometric problem-solving performance, leading a 21\% improvement over the previous state-of-the-art. 
\sciapproach surpasses \geoapproach on 6 of 7 tasks, indicating the function-driven mathematical knowledge strengthens cross-domain deductive reasoning.
Due to the broad-spectrum reasoning capacity afforded by injecting core knowledge of mathematical functions, \geoapproach and \sciapproach demonstrate significant academic merit and considerable practical potential for deployment in real-world intelligent education scenarios.

\begin{table*}[!htb]
    \centering
    \resizebox{0.98\linewidth}{!}{
    \begin{tabular}{lccccccc}
        \toprule
        \multicolumn{1}{l}{\multirow{2}{*}[-1.5ex]{\textbf{Models}}} & \textbf{MathVista} & \multicolumn{6}{c}{\textbf{MMMU}} \\ 
        \cmidrule(lr){2-2} \cmidrule(lr){3-8}
        & \makecell{\textbf{Geometry}} \raisebox{-0.3\height}{\includegraphics[height=0.6cm]{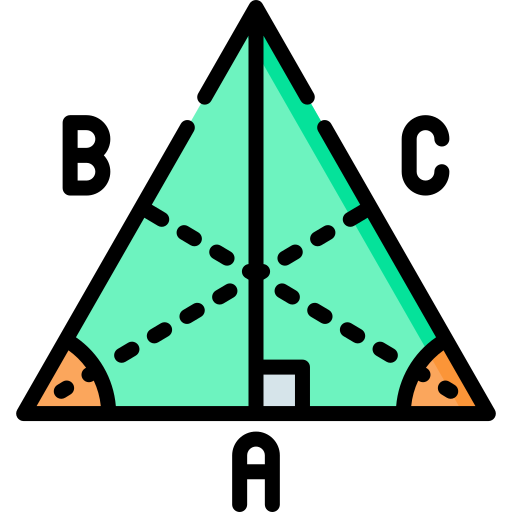}}
        & \makecell{\textbf{Art \&} \\ \textbf{Design}} \raisebox{-0.3\height}{\includegraphics[height=0.6cm]{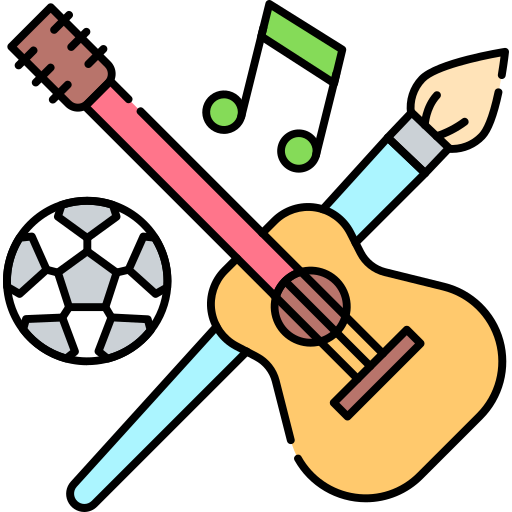}}
        & \makecell{\textbf{Business} \\} \raisebox{-0.3\height}{\includegraphics[height=0.6cm]{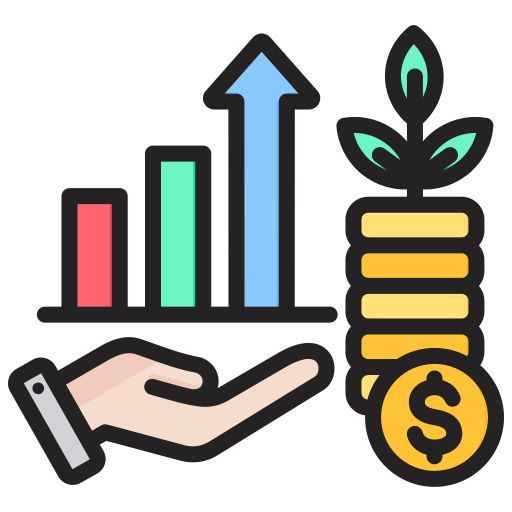}}
        & \makecell{\textbf{Science} \\} \raisebox{-0.3\height}{\includegraphics[height=0.6cm]{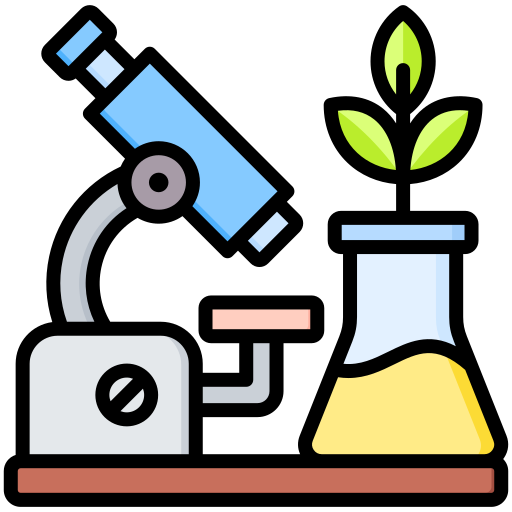}}
        & \makecell{\textbf{Health \&} \\ \textbf{Medicine}} \raisebox{-0.3\height}{\includegraphics[height=0.6cm]{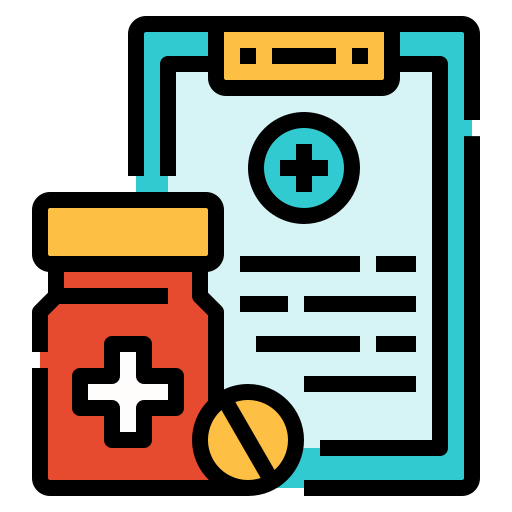}}
        & \makecell{\textbf{Human. \&} \\ \textbf{Social Sci.}} \raisebox{-0.3\height}{\includegraphics[height=0.6cm]{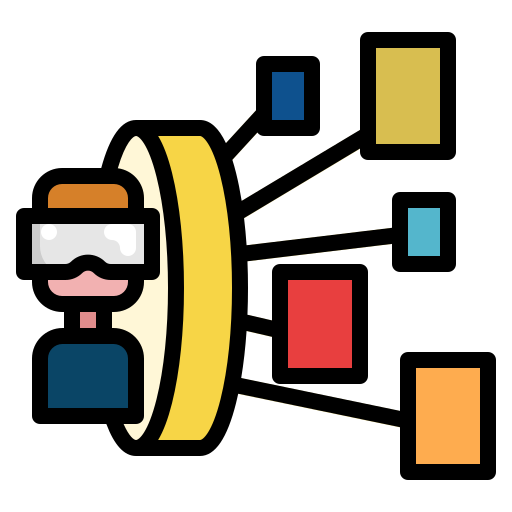}}
        & \makecell{\textbf{Tech. \&} \\ \textbf{Eng.}} \raisebox{-0.3\height}{\includegraphics[height=0.6cm]{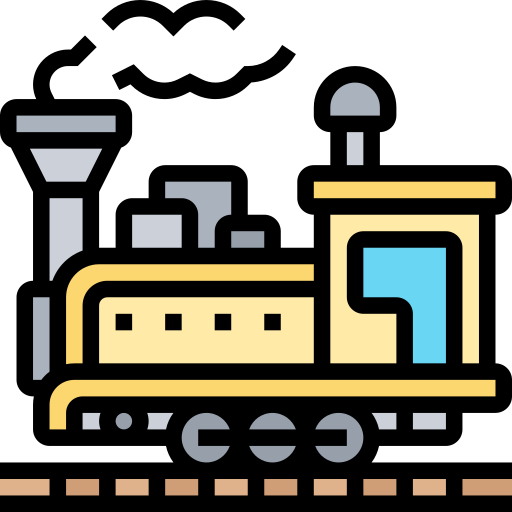}} \\
        \midrule
        Gemma3 & $60.7{\pm}1.9$ & $57.8{\pm}4.0$ & $44.1{\pm}0.6$ & $34.3{\pm}0.9$ & $46.8{\pm}2.2$ & $59.2{\pm}2.1$ & $29.0{\pm}1.3$ \\
        AutoGeo & $62.3{\pm}2.4$ & $59.3{\pm}1.4$ & $43.3{\pm}1.1$ & $34.9{\pm}1.3$ & $47.4{\pm}1.1$ & $58.9{\pm}1.5$ & $30.7{\pm}2.6$ \\
        GeoPeP & $61.0{\pm}2.3$ & $59.2{\pm}1.1$ & $40.4{\pm}1.4$ & $34.0{\pm}1.7$ & $45.1{\pm}0.9$ & $59.6{\pm}1.0$ & $32.6{\pm}0.7$ \\
        GeoGPT4V & $60.5{\pm}0.7$ & $60.2{\pm}1.1$ & $43.1{\pm}1.5$ & $34.5{\pm}0.7$ & $46.0{\pm}0.7$ & $58.3{\pm}1.6$ & $30.8{\pm}2.0$ \\
        Geo170K & $62.2{\pm}1.5$ & $58.5{\pm}0.8$ & $43.6{\pm}1.4$ & $30.9{\pm}2.0$ & $46.8{\pm}2.2$ & $59.9{\pm}2.2$ & $30.9{\pm}1.6$ \\
        GeoReasoning & $62.8{\pm}1.3$ & $60.2{\pm}2.0$ & $44.5{\pm}2.5$ & $36.0{\pm}2.0$ & $46.7{\pm}1.1$ & $60.0{\pm}0.5$ & $32.9{\pm}1.3$ \\
        \textbf{\geoapproach} 
        & \colorbox{backblue!75}{$75.9{\pm}0.8$}
        & \colorbox{backblue!75}{$64.0{\pm}1.6$}
        & \colorbox{backblue!75}{$50.5{\pm}1.9$}
        & \colorbox{backblue!75}{$42.2{\pm}0.6$}
        & \colorbox{backblue!75}{$48.4{\pm}1.8$}
        & \colorbox{backblue!75}{$64.7{\pm}0.8$} 
        & \colorbox{backred!50}{$37.6{\pm}1.4$} \\
        \textbf{\sciapproach} 
        & \colorbox{backred!50}{$81.8{\pm}1.1$} 
        & \colorbox{backred!50}{$65.7{\pm}1.2$}
        & \colorbox{backred!50}{$51.3{\pm}2.0$} 
        & \colorbox{backred!50}{$48.3{\pm}1.0$} 
        & \colorbox{backred!50}{$49.5{\pm}1.3$} 
        & \colorbox{backred!50}{$66.6{\pm}0.4$} 
        & \colorbox{backblue!75}{$35.2{\pm}1.1$} \\
        \bottomrule
    \end{tabular}}
    \caption{\textbf{Generalized deliberate reasoning Performance}. The abbreviations ``Human. \& Social Sci.'' and ``Tech. \& Eng.'' refer to ``Humanities and Social Science'' and ``Technology and Engineering''. 
    }
    \label{table4}
    \vspace{-1em}
\end{table*}

\begin{figure*}[!t]
    \centering
    \includegraphics[width=1.0\textwidth]{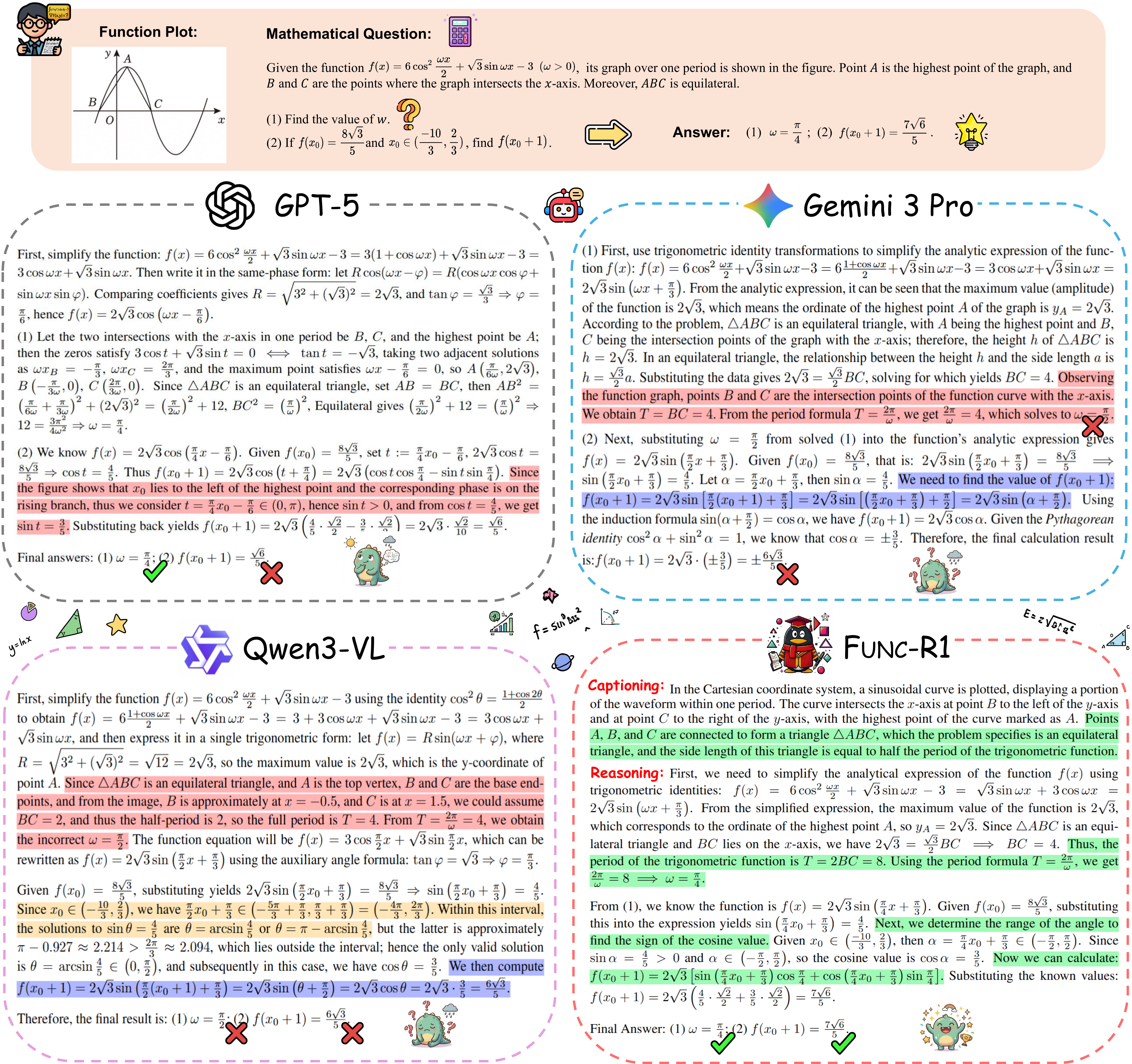}
    \caption{
    Qualitative evaluation of deductive reasoning in mathematical function problems.
    The \hlgreen{\textbf{precise solving procedure}} is highlighted, and we delineate discrete categories of causative factors that culminated in erroneous solutions: \hlred{\textbf{inaccurate visual perception}}, \hlpurple{\textbf{theoretical knowledge deficit}}, and \hlyellow{\textbf{defective logical reasoning}}.}
    \label{figure5}
    \vspace{-1em}
\end{figure*}

\subsection{Qualitative Reasoning Analysis}
We also provide a qualitative analysis to demonstrate practical utility of the constructed model. 
As shown in Figure~\ref{figure5}, when confronted with complex function-solving problems, GPT-5, Gemini 3 Pro, and Qwen3-VL demonstrate varying degrees of visual perceptual and logical reasoning hallucinations, leading to suboptimal solutions. 
In contrast, empowered by the hierarchical training regimen, \ourapproach explicitly decouples the function-solving pipeline and accomplishes rigorous mathematical reasoning grounded in meticulous visual interpretation.
This demonstrates that \ourapproach is a versatile AI assistant, coupling numerically robust symbolic derivation with automated reasoning over mathematical functions, enabling reliable decision support in engineering design, quantitative finance, and scientific computing.

\subsection{Ablation Studies}
\label{ablation_study}
\begin{figure}[htbp]
    \centering
    \includegraphics[width=\linewidth]{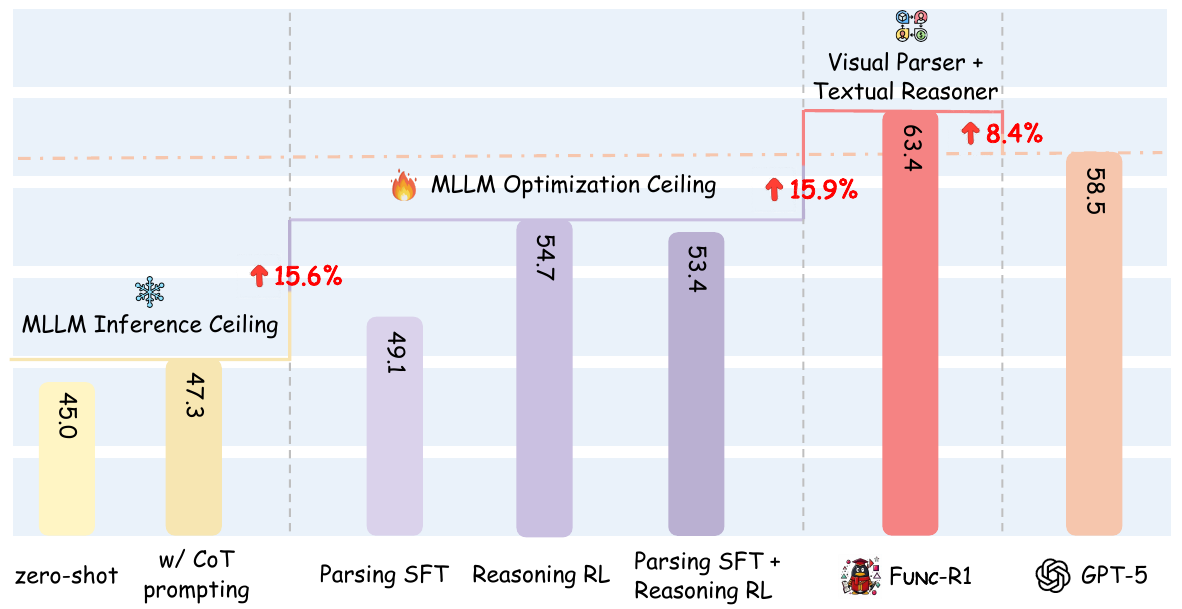}
    \caption{Reasoning bottlenecks in single-MLLMs.}
    \label{MLLM_bottlenecks}
\end{figure}

\textbf{Single-MLLM Bottlenecks.}
Herein, we leverage Qwen3-VL-8B to investigate the upper bound of reasoning performance achievable by a single MLLM in MathVerse. We comprehensively apply CoT prompting, SFT, and RL strategies to improve the MLLM’s abilities in computational reasoning. Figure~\ref{MLLM_bottlenecks} reveals the performance ceiling of a single open-source MLLM still remains substantially below that of GPT-5, achieving 54.7 compared to GPT-5's 58.5. \ourapproach employs a decoupled two-stage problem-solving pipeline that mitigates the performance bottleneck of a single MLLM architecture, surpassing GPT-5 by 8.4\% in mathematical reasoning performance.

\begin{figure}[htbp]
    \centering
    \includegraphics[width=\linewidth]{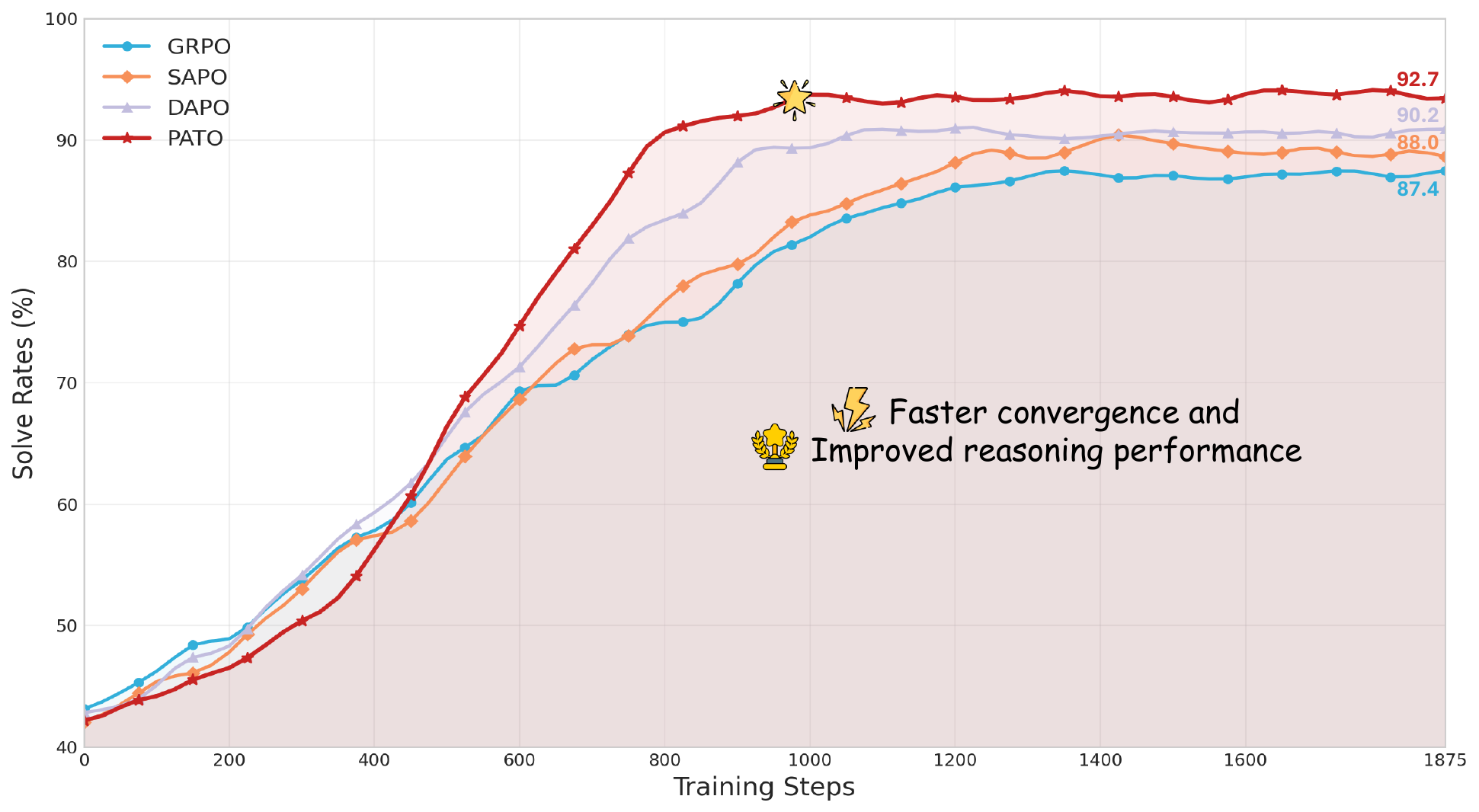}
    \caption{Policy optimization algorithm comparison.}
    \label{policy_updating}
\end{figure}

\textbf{Policy Gradient Updates.}
During the RL training phase, we replace hard gradient clipping in GRPO with a smooth, temperature-controlled trust region and incorporate comprehensive reward signals, thereby enabling more stable and efficient policy optimization. 
We conduct a comprehensive comparison among widely used gradient update algorithms, with the reward function, decoding configuration, and evaluation protocol kept identical. On the \Reasoning validation set, Figure~\ref{policy_updating} demonstrates that PATO converges faster and achieves superior solve rates compared to vanilla GRPO, SAPO, and DAPO.

\textbf{Reward Signal Design.}
The customized reward signals are calibrated to specifically address function-oriented reasoning challenges. 
We present comprehensive ablation results under full and partial reward signals in Table~\ref{ablation_rewards}.
Metrics indicate that the absence of any reward leads to degraded performance.
$\mathcal{R}_{\operatorname{s}}$ serves as the cornerstone for reasoning optimization, while $\mathcal{R}_{\operatorname{v}}$ and $\mathcal{R}_{\operatorname{k}}$ further advance visual perception correction and theoretical knowledge assimilation.

\begin{table}[!htb]
    \centering
    \resizebox{0.95\columnwidth}{!}{
    \begin{tabular}{ccc|ccc}
        \toprule
        \multicolumn{3}{c|}{\textbf{Rewards}} & \multirow{2}{*}{\textbf{MathVista}} & \multirow{2}{*}{\textbf{Math-Vision}} & \multirow{2}{*}{\textbf{MathVerse}} \\
        $\mathcal{R}_{\operatorname{s}}$ &
        $\mathcal{R}_{\operatorname{v}}$ & 
        $\mathcal{R}_{\operatorname{k}}$ & & & \\
        \midrule
        \textcolor{ForestGreen}{\usym{2713}} & & 
        & 65.7 & 32.1 & 51.4 \\
        \textcolor{ForestGreen}{\usym{2713}} & \textcolor{red}{\usym{2717}} & \textcolor{ForestGreen}{\usym{2713}}
        & 69.3 & 36.0 & 54.6 \\
        \textcolor{ForestGreen}{\usym{2713}} & \textcolor{ForestGreen}{\usym{2713}} & \textcolor{red}{\usym{2717}} 
        & 70.5 & 37.4 & 53.2 \\
        \textcolor{ForestGreen}{\usym{2713}} & \textcolor{ForestGreen}{\usym{2713}} & \textcolor{ForestGreen}{\usym{2713}} 
        & \colorbox{backred!50}{73.7} 
        & \colorbox{backred!50}{42.4} 
        & \colorbox{backred!50}{57.5} \\
        \bottomrule
    \end{tabular}}
    \caption{Ablation study of reward signal components.}
    \label{ablation_rewards}
\end{table}

\section{Related Work}
\subsection{Multimodal Mathematical Benchmarks}
Rigorous assessment of the mathematical reasoning capabilities of multimodal foundation models requires comprehensive and systematic mathematical benchmarks. MMMU~\citep{MMMU}, MathVista~\citep{MathVista}, and Math-Vision~\citep{Math-Vision} innovatively incorporate numerous multimodal mathematical problems to comprehensively evaluate the reasoning performance. 
Through an in-depth evaluation of mathematical reasoning, MathVerse~\citep{MathVerse} reveals persistent limitations in MLLMs’ diagram perception. 
LiveK12Bench~\citep{LiveK12Bench} targets key problem-solving challenges arising from exam parsing, process verification, and efficiency constraints. 
Additionally, We-Math~\citep{We-Math} explores the progressive mathematical reasoning principles beyond holistic performance. 

\subsection{Mathematical Reasoning Optimization}
Accurate mathematical problem-solving in multimodal scenarios depends on the integration of robust visual perception with meticulous mathematical reasoning. Substantial efforts have been devoted to monolithically improve MLLMs’ mathematical problem-solving capabilities. 
Specifically, Math-LLaVA~\citep{Math-LLaVA}, MathCoder-VL~\citep{MathCoder-VL}, G-LLaVA~\citep{G-LLaVA}, and Math-LMM~\cite{CMM-Math} employ instruction tuning to enhance the cross-modal alignment between visual diagrams and textual content.
Furthermore, Vision-R1~\citep{Vision-R1}
leverage reinforcement learning techniques to advance stepwise mathematical problem-solving.

\section{Conclusion}
This work takes mathematical function reasoning as a testbed to reveal that contemporary MLLMs still exhibit a fundamental tension between precise visual perception and rigorous intricate reasoning. 
The curated \Caption and \Reasoning corpora constitute robust empirical resources that underpin an emerging research direction focused on mathematical function problem-solving under multimodal reasoning scenarios.
We present \ourapproach, a multimodal reasoning LLM with advanced mathematical proficiency, which delivers sophisticated reasoning grounded in profound visual parsing.
The proposed hierarchical training strategy integrates domain-specific supervised fine-tuning and perception-aligned theoretically grounded policy optimization, furnishing broad guidance for future multimodal reasoning optimization studies.
\ourapproach attains outstanding sophisticated function reasoning capability commensurate with GPT-5, and the proposed techniques effectively foster cross-disciplinary generalized reasoning.

\section*{Limitations}
In this research work, the amount of high-quality data for function-oriented multimodal mathematical reasoning remains relatively limited. Building such data requires careful alignment among graph perception, mathematical property annotation, and solution-level reasoning, which makes large-scale collection costly and challenging. Although our current dataset is sufficient to support the main findings, broader coverage of function types, visual forms, and problem difficulty would likely further improve robustness and generalization. 
Moreover, while the decoupled architecture design helps alleviate modality interference, it may still suffer from error propagation from visual parsing to downstream reasoning. Future work will further explore more interactive perception-reasoning mechanisms to enhance reliability across diverse multimodal reasoning settings.

\section*{Acknowledgments}
This research is supported by the National Natural Science Foundation of China under Grant No. U22A6001 and No. 62606436.


\bibliography{references}

\appendix

\section{Large Language Model Usage}
In this research, large language models are utilized as generation and proofreading tools to produce mathematical function reasoning data and refine academic English expression. The model development relies exclusively on publicly available, open-source foundation models from the Qwen series~\citep{Qwen3, Qwen3-VL}.
It is worth noting that the conceptual innovation and motivation underlying the study are conceived independently by the human author team, without any conceptual or cognitive inspiration from large language models. 
Multiple large language models, including GPT-5~\citep{GPT-5}, Gemini 3 Pro~\citep{Gemini-3-Pro}, Claude Opus 4.5~\citep{Claude-4.5}, QWen3-VL-235B-A22B-Thinking~\citep{Qwen3-VL}, and GLM-4.1V-9B-Thinking~\citep{GLM-4.1V-Thinking}, are used within the data generation engine to annotate function plots and design reasoning-oriented problem solving data. We employ large language models exclusively through the official API endpoints provided by respective vendors. All human–AI interactions are conducted in full compliance with all applicable terms of service and licensing conditions. We solemnly affirm that the constructed dataset is used exclusively for academic research and entails no commercial purpose.
The manuscript is originally written by the human authors, and large language models are engaged only at the final stage to assist in polishing key sections of the language and word choice.

\section{Ethical Considerations}
\begin{itemize}
  \item \textbf{Copyright and Licensing:} Adherence to all applicable copyright and licensing requirements is mandatory.
  \item \textbf{Data Privacy:} Compliance with data protection laws and ethical standards in data handling is paramount. Annotators must refrain from collecting questions or content containing personal or sensitive information.
  \item \textbf{Model Development:} The developed model is built upon publicly available pre-trained Qwen series foundation models~\citep{Qwen3, Qwen3-VL}. We disclose the exact model versions, licensing terms, and training details, and we comply with all license terms and usage restrictions.
\end{itemize}

\section{Reproducibility Statement}
\label{implement_details}
To alleviate modality interference in mathematical function reasoning, we explicitly decouple the architecture into two specialized components: a visual math parser and a textual math reasoner. An ablation study examining the performance ceiling of a unified multimodal architecture is presented in Sec. 3.6, Single-MLLM Bottlenecks.
Specifically, \ourapproach-Lite utilizes Qwen3-VL-4B as the visual math parser, and adopts Qwen3-8B to serve as the textual math reasoner for logical reasoning computation. 
\ourapproach-Pro employs Qwen3-VL-8B as the visual math parser and Qwen3-32B as the textual math reasoner.
For the visual math parser, the SFT stage runs for 3 epochs with batch size 8 and learning rate $1e{-5}$. 
For the textual math reasoner, the RL stage lasts for 5 epochs with batch size 64, learning rate $1e{-6}$, and a rollout number $N=8$.



\section{Detailed Prompt Template}

\begin{promptbox}{Problem-Solving Prompt}
\small
You are a mathematics expert. Your task is to combine the given function expression with the attached graph, carry out rigorous mathematical reasoning, and provide a complete solution. In your reasoning and calculations, explicitly use standard theoretical properties of mathematical function. 
Present your step-by-step solution process and mark your final answer according to the following rules: \\
- Single-choice: place the correct option letter inside \verb|\boxed{}|. \\
- Multiple-choice: place all correct option letters inside one \verb|\boxed{}|. \\
- Fill-in-the-blank: enclose each blank’s answer in a separate \verb|\boxed{}|. \\
- Open-ended: enclose each sub-question’s answer in a separate \verb|\boxed{}|.
\end{promptbox}

\begin{promptbox}{Answer Extraction Prompt}
\small
You are an expert grader. Determine whether the student's answer is consistent with the standard answer. 
In accordance with the following grading criteria: \\
- Judge each sub-question independently. \\
- Accept equivalent expressions (mathematical or textual). If equivalence cannot be determined, mark the answer incorrect. \\
- Do not re-derive the answer; for questions with definitive results, compare the final answer only. For non-definitive questions (\textit{e.g.}, proofs), evaluate whether the reasoning approach is sound. \\
- Output the number of correctly answered sub-questions, enclosed in \verb|\boxed{}|.
\end{promptbox}









\section{Additional Experimental Details}
\label{experiment_details}
\textbf{LLM baselines.} 
To facilitate comprehensive comparison, we diversely incorporate the current state-of-the-art open-source and closed-source MLLMs. 
Particularly, we select LLaVA series~\citep{LLaVA-NeXT}, InternVL series~\citep{InternVL3.5}, Qwen-VL series~\citep{Qwen3-VL}, DeepSeek-VL2~\citep{DeepSeek-VL2}, GLM-4.1V-Thinking~\citep{GLM-4.1V-Thinking}, CogVLM2~\citep{CogVLM2}, WeThink~\citep{WeThink}, R1-VL~\citep{R1-VL}, and Math-LLaVA~\citep{Math-LLaVA} for open-source models. 
Additionally, for closed-source models, we employ Qwen-VL-Max~\citep{Qwen-VL-Max}, Gemini 2.5 Flash, Gemini 3 Flash~\citep{Gemini-3-Pro}, Claude Opus 4.1~\citep{Claude-4.1}, GPT-4o, GPT-4V~\citep{GPT-4}, and GPT-5~\citep{GPT-5}.

\textbf{Evaluation benchmarks.}
Experiments are holistically conducted on in-distribution benchmark \Reasoning, as well as diverse out-of-distribution multimodal mathematical reasoning benchmarks, including MathVista~\citep{MathVista}, Math-Vision~\citep{Math-Vision}, MathVerse~\citep{MathVerse}, and We-Math~\citep{We-Math}.
We consistently employ an open-ended evaluation protocol whereby the final answer is extracted from the LLM-based solver’s output and its correctness is established through comparison with the ground-truth solution.
Implementation details are provided in the Appendix~\ref{implement_details}.

\section{Data Generative Design Principles}
When executing the knowledge concept sedimentation, LLMs (\textit{i.e.}, GPT-5~\citep{GPT-5}, Gemini 3 Pro~\citep{Gemini-3-Pro}) receive the sample-level function annotations and the knowledge concept panel from the previous iteration. Instead of relying on downstream reasoning feedback, the thinkers directly optimize the knowledge content based on the three following steps: 

(i) \textit{Parsing:} Thinker parses the sample-level concept into a structured key-value format. 

(ii) \textit{Optimization:} Thinker scrutinizes the extracted knowledge concepts against the fundamental requirements. In the initial round, it directly condenses raw descriptions; in subsequent rounds, it filters out superficial, redundant, or misleading keys that fail to reflect intrinsic mathematical properties.

(iii) \textit{Integration:} The remaining high-value knowledge items are aggregated into a consolidated knowledge concept panel, marking the completion of the current iteration. 

To ensure rich coverage and high-quality design throughout the self-evolving knowledge consolidation engine, we construct mathematical function-oriented question-answer pairs following three fundamental requirements: 

(i) \textit{Logical Reachability}: The questions must be framed within the constraints of the given function plots and captions, allowing for final answers to be reached logically.

(ii) \textit{Intrinsic Focus}: Generative LLMs is guided to focus on the intrinsic properties of mathematical functions, avoiding dependence on superficial visual cues.

(iii) \textit{Concept Assessment}: Designed function problems tend to assess reasoning abilities related to at least one relevant knowledge concept.

This engine offers a rigorous, reliable pipeline for generating high-quality multimodal mathematical reasoning data. The resulting \Caption and \Reasoning comprise 3,519 entries, providing a solid foundation for downstream multimodal mathematical function reasoning applications.


\section{Empirical Analysis of Mathematical Functions}

Herein, we elucidate our research motivation by highlighting the unique challenges of function reasoning and the necessity of task-specific reward signals. 
The mathematical function’s explicit parameterization, computational complexity, and robust measurability pose reasoning obstacles that diverge from traditional problem formats.
We conduct a rigorous, systematic comparative analysis to characterize GPT-5’s problem-solving error distribution across Multiple problem categories from MathVerse (\textit{i.e.}, plane geometry, solid geometry, mathematical function). We also integrate the novel \Reasoning dataset into the analysis, and detailed composition statistics are shown in Table~\ref{table1}. 

\begin{table}[!ht]
    \centering
    \resizebox{0.98\linewidth}{!}{
    \begin{tabular}{lr}
        \toprule
        \textbf{Problem Category} & \textbf{Number \& Percentage} \\
        \midrule
        MathVerse~\citep{MathVerse} & \\
        \quad - Plane geometry \raisebox{-0.3\height}{\includegraphics[height=0.6cm]{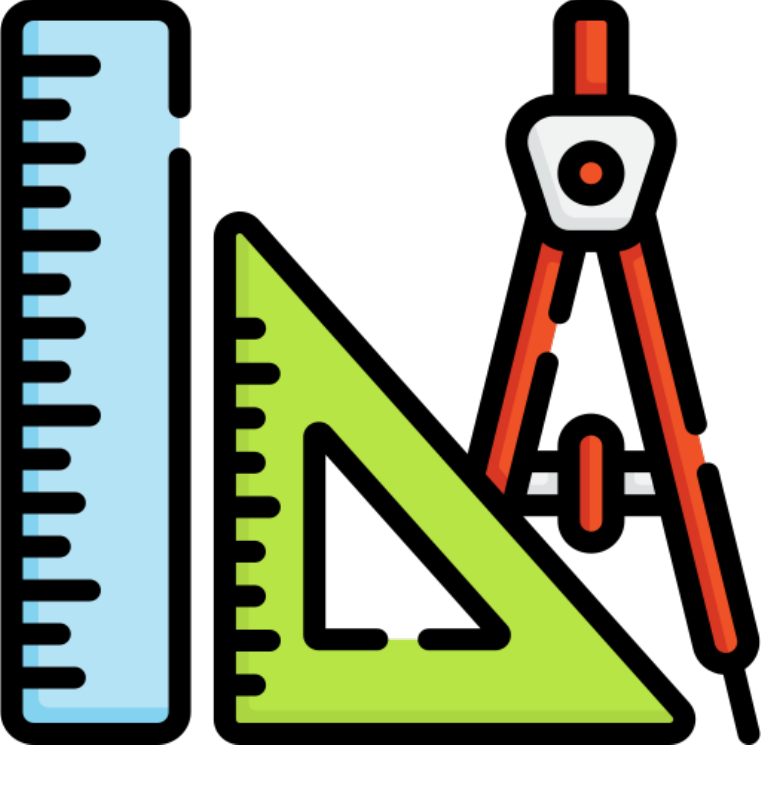}} & 1,746 (66.8\%)  \\
        \quad - Solid geometry \raisebox{-0.3\height}{\includegraphics[height=0.65cm]{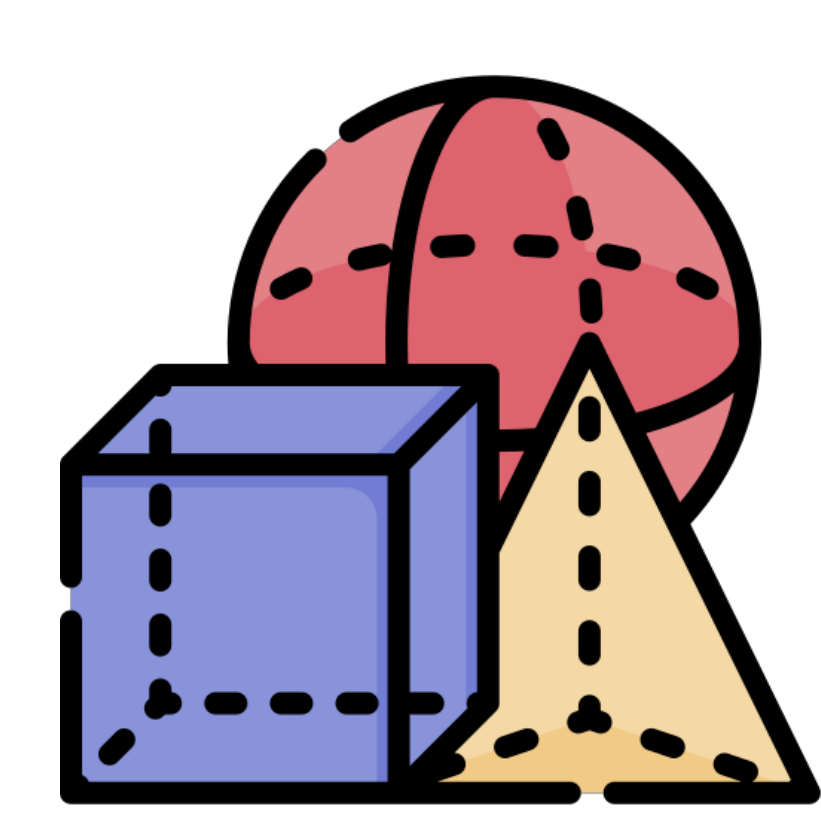}} & 332 (12.7\%) \\
        \quad - \textbf{Mathematical function} \raisebox{-0.3\height}{\includegraphics[height=0.55cm]{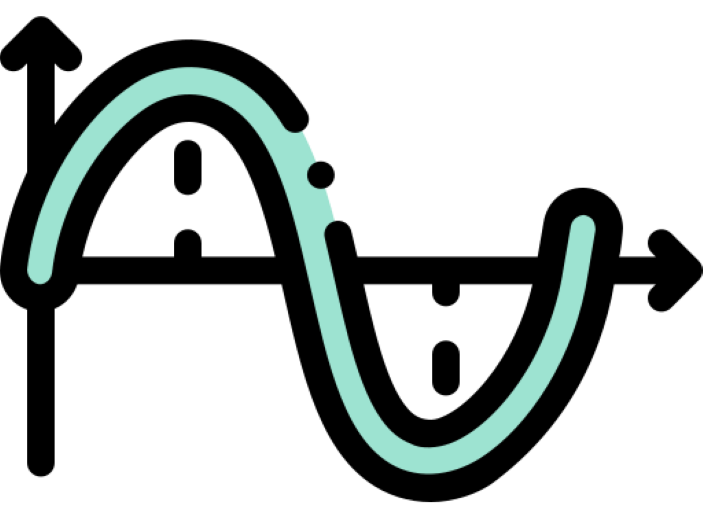}} & \textbf{534 (20.5\%)} \\
        \midrule
        \Reasoning & \\
        \quad - \textbf{Mathematical function} \raisebox{-0.3\height}{\includegraphics[height=0.55cm]{function.png}} & \textbf{3,519 (100\%)} \\
        \bottomrule
    \end{tabular}}
    \caption{Key statistics of the mathematical database.}
    \label{table1}
\end{table}

\begin{figure}[!ht]
    \centering
    \includegraphics[width=0.98\linewidth]{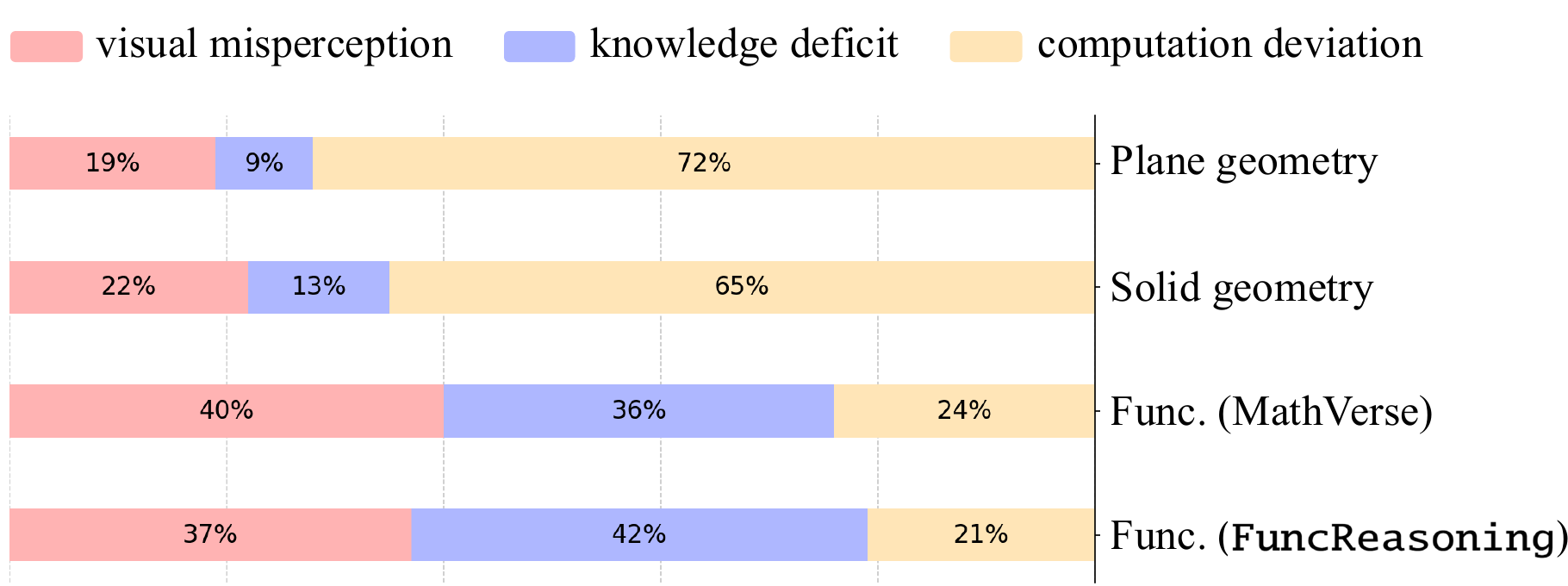}
    \caption{Attribution distribution of mathematical problem-solving errors.}
    \label{figure3}
\end{figure}

A comprehensive depiction of error-type proportions for distinct problem categories is shown in Figure~\ref{figure3}. 
With increasing problem difficulty (\textit{i.e.}, from plane geometry through solid geometry to function), the predominant error types transition from computation deviation to visual misperception and knowledge deficit. This further reinforces the limitations of existing approaches that optimize exclusively for rigorous computational reasoning.
As detailed in Sec.~\ref{reward_signal_design}, we tailored comprehensive reward signals within the PATO strategy (\textit{i.e.}, visual perception correction, theoretical knowledge guidance, and verifiable problem-solving) to deliver targeted interventions for challenges encountered in function-oriented complex reasoning.

\begin{table*}[!ht]
\centering

\tcbset{
  apiCard/.style={
    colback=white,
    colframe=blue!60!black,    
    colbacktitle=blue!10!white, 
    coltitle=black,
    boxrule=0.8pt,
    arc=1mm,
    width=\textwidth,
    left=6pt, right=6pt, top=2pt, bottom=2pt, 
    fonttitle=\bfseries\sffamily,
    title=#1,
    after skip=0.04cm                         
  }
}

\tcbset{
  openCard/.style={
    colback=white,
    colframe=green!60!black,   
    colbacktitle=green!10!white,
    coltitle=black,
    boxrule=0.8pt,
    arc=1mm,
    width=\textwidth,
    left=6pt, right=6pt, top=2pt, bottom=2pt,
    fonttitle=\bfseries\sffamily,
    title=#1,
    after skip=0.04cm
  }
}


\begin{tcolorbox}[openCard={LLaVA series}~\citep{LLaVA-NeXT}]
  By leveraging high-quality multi-turn visual instruction data and synthetic augmentation, LLaVA scales effectively and improves performance on complex visual question answering, fine-grained image description, and multimodal reasoning. This framework features a lightweight, efficient training pipeline and an open data/code ecosystem that facilitate rapid iteration.
\end{tcolorbox}

\begin{tcolorbox}[openCard={InternVL series}~\citep{InternVL3.5}]
  A family of general-purpose vision-language models for industrial and research applications, emphasizing large-scale pretraining and fine-grained region alignment. 
  Integrating high-resolution feature aggregation with dynamic receptive fields, the models enhance dense-scene and OCR-intensive comprehension and enable robust long-context multi-image reasoning and cross-page document QA.
\end{tcolorbox}

\begin{tcolorbox}[openCard={CogVLM2}~\citep{CogVLM2}]
  CogVLM2 emphasizes stronger visual encoding and language alignment. Trained with larger-scale multimodal instruction data and high-resolution inputs, it exhibits robustness on web page screenshots, GUIs, charts, and scene text.
\end{tcolorbox}

\begin{tcolorbox}[openCard={Qwen-VL series}~\citep{Qwen3-VL}]
  Developed by the Qwen multimodal team, this system integrates strong Chinese language proficiency with deep visual understanding, focusing on OCR, chart analysis, practical QA, and tool use across mobile and server deployments. It leverages high-resolution image encoding and region-level supervision, supports multi-image, multi-turn, and multi-format inputs.
\end{tcolorbox}

\begin{tcolorbox}[openCard={DeepSeek-VL2}~\citep{DeepSeek-VL2}]
  DeepSeek-VL2 prioritizes efficient alignment and robust reasoning. Through refined visual pretraining and instruction distillation, it reduces hallucinations and improves complex QA performance.
\end{tcolorbox}

\begin{tcolorbox}[openCard={GLM-4.1V-Thinking}~\citep{GLM-4.1V-Thinking}]
  The thinking variant of the GLM-4 series emphasizes explicit multi-step chain-of-thought and visual reasoning, supporting coordinated image–text planning and intermediate reasoning traces to enhance interpretability on complex tasks.
\end{tcolorbox}

\vspace{0.1cm}


\begin{tcolorbox}[apiCard={Math-LLaVA}~\citep{Math-LLaVA}]
  Strengthened by math-domain instruction data and structured annotations, Math-LLaVA improves symbol recognition, quantitative derivation, while improving visual–semantic alignment to reduce formula recognition errors.
\end{tcolorbox}

\begin{tcolorbox}[apiCard={WeThink}~\citep{WeThink}]
  By integrating process rewards and preference alignment, WeThink improves robustness in visual reasoning and decision-making. The framework supports multi-image, multi-evidence integration to mitigate hallucination and inconsistency.
\end{tcolorbox}

\begin{tcolorbox}[apiCard={R1-VL}~\citep{R1-VL}]
  R1-VL emphasizes stepwise reasoning, error correction, and reflective alignment to improve performance on long-horizon tasks. The approach yields notable gains on geometric math, scientific chart interpretation, and quantitative visual analysis.
\end{tcolorbox}

\vspace{1em}
\caption{Detailed model introduction card for incorporated baselines.}
\label{tab:model_card}
\end{table*}

\section{Supplementary Baseline Details}
\label{baseline_details}
Additional introductory details on the included baseline models are provided in the form of model cards, as illustrated in Table~\ref{tab:model_card}.

\section{Mathematical Reasoning Examples}
\begin{figure*}[b]
    \centering
    
    \begin{subfigure}{\textwidth}
        \centering
        \includegraphics[width=\textwidth, height=0.25\textheight, keepaspectratio]{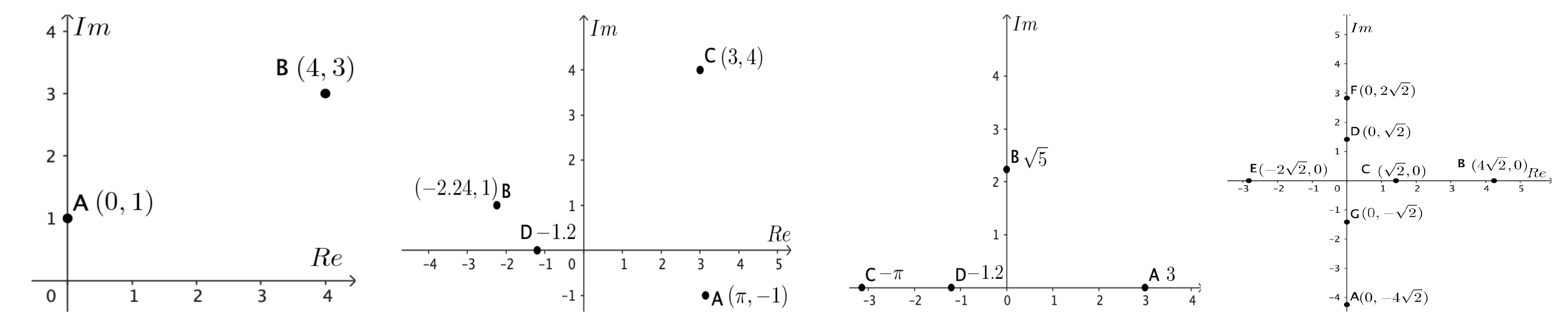}
        \caption{Examples of the mathematical diagram for discrete function.}
    \end{subfigure}
    
    \vspace{1.5cm}
    
    \begin{subfigure}{\textwidth}
        \centering
        \includegraphics[width=\textwidth, height=0.25\textheight, keepaspectratio]{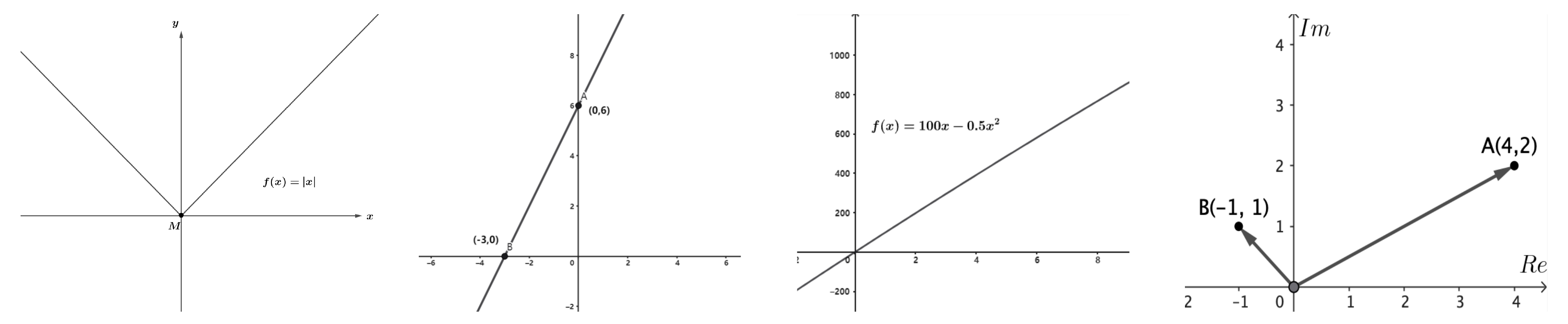}
        \caption{Examples of the mathematical diagram for linear function.}
    \end{subfigure}
    
    \vspace{1.5cm}
    
    \begin{subfigure}{\textwidth}
        \centering
        \includegraphics[width=\textwidth, height=0.25\textheight, keepaspectratio]{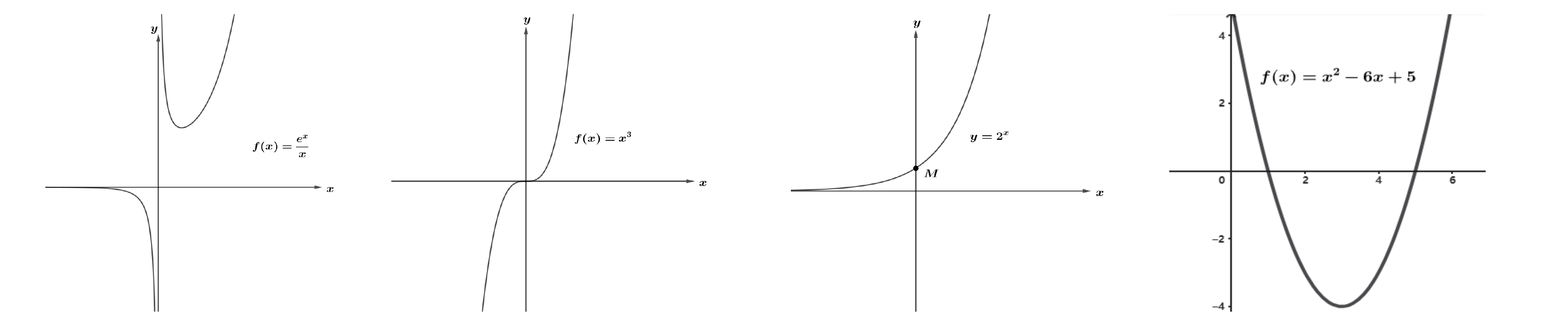}
        \caption{Examples of the mathematical diagram for nonlinear function.}
    \end{subfigure}
    
    \vspace{1.5cm}
    
    \begin{subfigure}{\textwidth}
        \centering
        \includegraphics[width=\textwidth, height=0.25\textheight, keepaspectratio]{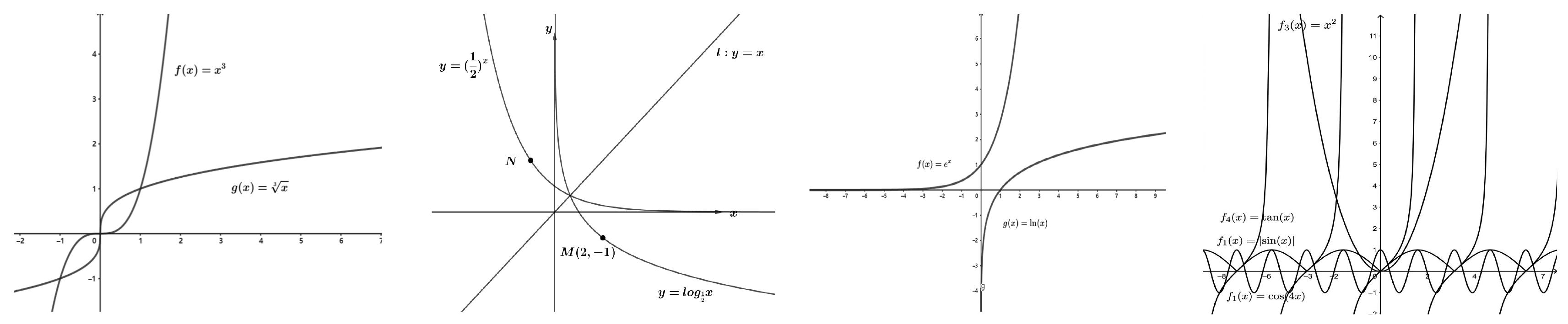}
        \caption{Examples of the mathematical diagram for composite function.}
    \end{subfigure}

    \vspace{0.5cm}
    
\caption{Function-oriented mathematical visualization of Synthetic Datasets.}
\label{figure:data_examples}
\end{figure*}

We assemble \Caption and \Reasoning as comprehensive corpora as possible of diverse function diagrams to facilitate \ourapproach’s thorough acquisition of foundational mathematical theoretical knowledge. As illustrated in Figure~\ref{figure:data_examples}, we provide a detailed presentation of high-quality mathematical visualizations that encompass the discrete, linear, nonlinear, and composite functions.

\begin{table*}[p]
    \centering
    \begin{tabular}{p{0.95\textwidth}}
        \hline
        \multicolumn{1}{c}{
        \raisebox{-0.22cm}{\includegraphics[width=0.65cm, height=0.65cm]{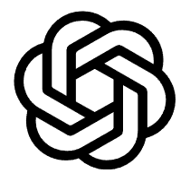}} \
        \rule{0pt}{3ex}\textbf{GPT-5} 
        \rule[-2ex]{0pt}{0pt}}\\
        \hline
        \rule{0pt}{3ex}
        First, simplify the function: $f(x)=6\cos^2\frac{\omega x}{2}+\sqrt3\sin\omega x-3=3(1+\cos\omega x)+\sqrt3\sin\omega x-3=3\cos\omega x+\sqrt3\sin\omega x$. Then write it in the same-phase form: let $R\cos(\omega x-\varphi)=R(\cos\omega x\cos\varphi+\sin\omega x\sin\varphi)$. Comparing coefficients gives $R=\sqrt{3^2+(\sqrt3)^2}=2\sqrt3$, and $\tan\varphi=\frac{\sqrt3}{3}\Rightarrow\varphi=\frac{\pi}{6}$, hence $f(x)=2\sqrt3\cos\left(\omega x-\frac{\pi}{6}\right)$. 
        (1) Let the two intersections with the $x$-axis in one period be $B$, $C$, and the highest point be $A$; then the zeros satisfy $3\cos t+\sqrt3\sin t=0\iff\tan t=-\sqrt3$, taking two adjacent solutions as $\omega x_B=-\frac{\pi}{3}$, $\omega x_C=\frac{2\pi}{3}$, and the maximum point satisfies $\omega x-\frac{\pi}{6}=0$, so $A\left(\frac{\pi}{6\omega},2\sqrt3\right)$, $B\left(-\frac{\pi}{3\omega},0\right)$, $C\left(\frac{2\pi}{3\omega},0\right)$. Since $\triangle ABC$ is an equilateral triangle, set $AB=BC$, then $AB^2=\left(\frac{\pi}{6\omega}+\frac{\pi}{3\omega}\right)^2+(2\sqrt3)^2=\left(\frac{\pi}{2\omega}\right)^2+12$, $BC^2=\left(\frac{\pi}{\omega}\right)^2$, Equilateral gives $\left(\frac{\pi}{2\omega}\right)^2+12=\left(\frac{\pi}{\omega}\right)^2\Rightarrow 12=\frac{3\pi^2}{4\omega^2}\Rightarrow \omega=\frac{\pi}{4}$. 
        (2) We know $f(x)=2\sqrt3\cos\left(\frac{\pi}{4}x-\frac{\pi}{6}\right)$. Given $f(x_0)=\frac{8\sqrt3}{5}$, set $t:=\frac{\pi}{4}x_0-\frac{\pi}{6}$, $2\sqrt3\cos t=\frac{8\sqrt3}{5}\Rightarrow\cos t=\frac{4}{5}$. Thus $f(x_0+1)=2\sqrt3\cos\left(t+\frac{\pi}{4}\right)=2\sqrt3\left(\cos t\cos\frac{\pi}{4}-\sin t\sin\frac{\pi}{4}\right)$. Since the figure shows that $x_0$ lies to the left of the highest point and the corresponding phase is on the rising branch, thus we consider $t = \frac{\pi}{4}x_0-\frac{\pi}{6} \in(0,\pi)$, hence $\sin t>0$, and from $\cos t=\frac{4}{5}$, we get $\sin t=\frac{3}{5}$. Substituting back yields $f(x_0+1)=2\sqrt3\left(\frac{4}{5}\cdot\frac{\sqrt2}{2}-\frac{3}{5}\cdot\frac{\sqrt2}{2}\right)=2\sqrt3\cdot\frac{\sqrt2}{10}=\frac{\sqrt6}{5}$. 
        Final answers: (1) $\omega=\frac{\pi}{4}$; (2) $f(x_0+1)=\frac{\sqrt6}{5}$. 
        \rule[-2ex]{0pt}{0pt} \\
        \hline
        \multicolumn{1}{c}{
        \raisebox{-0.22cm}{\includegraphics[width=0.65cm, height=0.65cm]{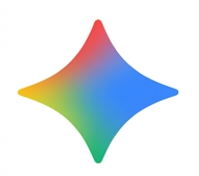}} \
        \rule{0pt}{3ex}\textbf{Gemini-3-Pro}
        \rule[-2ex]{0pt}{0pt}}\\
        \hline
        \rule{0pt}{3ex}
        (1) First, use trigonometric identity transformations to simplify the analytic expression of the function $f(x)$:
        $f(x)=6 \cos^2 \frac{\omega x}{2}+\sqrt{3} \sin \omega x-3 = 6 \frac{1+\cos \omega x}{2} + \sqrt{3} \sin \omega x - 3 = 3 \cos \omega x + \sqrt{3} \sin \omega x = 2\sqrt{3} \sin \left(\omega x+\frac{\pi}{3}\right)$.
        From the analytic expression, it can be seen that the maximum value (amplitude) of the function is $2\sqrt{3}$, which means the ordinate of the highest point $A$ of the graph is $y_A = 2\sqrt{3}$. According to the problem, $\triangle ABC$ is an equilateral triangle, with $A$ being the highest point and $B$, $C$ being the intersection points of the graph with the $x$-axis; therefore, the height $h$ of $\triangle ABC$ is $h = 2\sqrt{3}$. In an equilateral triangle, the relationship between the height $h$ and the side length $a$ is $h = \frac{\sqrt{3}}{2}a$. Substituting the data gives $2\sqrt{3} = \frac{\sqrt{3}}{2} BC$, solving for which yields $BC=4$. Observing the function graph, points $B$ and $C$ are the intersection points of the function curve with the $x$-axis. We obtain $T = BC = 4$. From the period formula $T = \frac{2\pi}{\omega}$, we get $\frac{2\pi}{\omega} = 4$, which solves to $\omega = \frac{\pi}{2}$.
        (2) Next, substituting $\omega = \frac{\pi}{2}$ from solved (1) into the function's analytic expression gives $f(x) = 2\sqrt{3} \sin \left(\frac{\pi}{2} x+\frac{\pi}{3}\right)$. Given $f(x_{0})=\frac{8\sqrt{3}}{5}$, that is:
        $2\sqrt{3} \sin \left(\frac{\pi}{2} x_{0}+\frac{\pi}{3}\right) = \frac{8\sqrt{3}}{5} \implies \sin \left(\frac{\pi}{2} x_{0}+\frac{\pi}{3}\right) = \frac{4}{5}$. Let $\alpha = \frac{\pi}{2} x_{0}+\frac{\pi}{3}$, then $\sin \alpha = \frac{4}{5}$. We need to find the value of $f(x_{0}+1)$:
        $f(x_{0}+1) = 2\sqrt{3} \sin \left[\frac{\pi}{2} (x_{0}+1)+\frac{\pi}{3}\right] = 2\sqrt{3} \sin \left[\left(\frac{\pi}{2} x_{0}+\frac{\pi}{3}\right) + \frac{\pi}{2}\right] = 2\sqrt{3} \sin \left(\alpha + \frac{\pi}{2}\right)$.
        Using the induction formula $\sin(\alpha + \frac{\pi}{2}) = \cos \alpha$, we have $f(x_{0}+1) = 2\sqrt{3} \cos \alpha$. Given the \textit{Pythagorean identity} $\cos^2 \alpha + \sin^2 \alpha = 1$, we know that $\cos \alpha  = \pm \frac{3}{5}$. Therefore, the final calculation result is:$f(x_{0}+1) = 2\sqrt{3} \cdot \left(\pm \frac{3}{5}\right) = \pm \frac{6\sqrt{3}}{5}$. 
        \rule[-2ex]{0pt}{0pt} \\
        \hline
    \end{tabular}
    \caption{The detailed solution provided by leading closed‑source models.}
\end{table*}

\begin{table*}[p]
    \centering
    \begin{tabular}{p{0.95\textwidth}}
        \hline
        \multicolumn{1}{c}{
        \raisebox{-0.22cm}{\includegraphics[width=0.65cm, height=0.65cm]{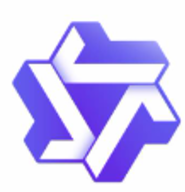}} \
        \rule{0pt}{3ex}\textbf{Qwen3-VL-235B-A22B} 
        \rule[-2ex]{0pt}{0pt}}\\
        \hline
        \rule{0pt}{3ex}
        First, simplify the function $ f(x) = 6 \cos^{2} \frac{\omega x}{2} + \sqrt{3} \sin \omega x - 3 $ using the identity $ \cos^{2} \theta = \frac{1 + \cos 2\theta}{2} $ to obtain  
        $f(x) = 6 \frac{1 + \cos \omega x}{2} + \sqrt{3} \sin \omega x - 3 = 3 + 3 \cos \omega x + \sqrt{3} \sin \omega x - 3 = 3 \cos \omega x + \sqrt{3} \sin \omega x$, and then express it in a single trigonometric form: let $ f(x) = R \sin(\omega x + \varphi) $, where $ R = \sqrt{3^2 + (\sqrt{3})^2} = \sqrt{12} = 2\sqrt{3} $, so the maximum value is $ 2\sqrt{3} $, which is the y-coordinate of point $A$.  
        Since $ \triangle ABC $ is an equilateral triangle, and $A$ is the top vertex, $B$ and $C$ are the base endpoints, and from the image, $B$ is approximately at $ x = -0.5 $, and $C$ is at $ x = 1.5 $, we could assume $ BC = 2 $, and thus the half-period is 2, so the full period is $ T = 4 $. From $ T = \frac{2\pi}{\omega} = 4 $, we obtain the incorrect $ \omega = \frac{\pi}{2}$.  
        The function equation will be $ f(x) = 3 \cos \frac{\pi}{2} x + \sqrt{3} \sin \frac{\pi}{2} x $, which can be rewritten as $ f(x) = 2\sqrt{3} \sin\left( \frac{\pi}{2} x + \frac{\pi}{3} \right) $ using the auxiliary angle formula: $ \tan \varphi  = \sqrt{3} \Rightarrow \varphi = \frac{\pi}{3}$.  
        Given $ f(x_0) = \frac{8\sqrt{3}}{5} $, substituting yields  
        $2\sqrt{3} \sin\left( \frac{\pi}{2} x_0 + \frac{\pi}{3} \right) = \frac{8\sqrt{3}}{5} \Rightarrow \sin\left( \frac{\pi}{2} x_0 + \frac{\pi}{3} \right) = \frac{4}{5}$. Since $ x_0 \in \left( -\frac{10}{3}, \frac{2}{3} \right) $, we have $ \frac{\pi}{2} x_0 + \frac{\pi}{3} \in \left( -\frac{5\pi}{3} + \frac{\pi}{3}, \frac{\pi}{3} + \frac{\pi}{3} \right) = \left( -\frac{4\pi}{3}, \frac{2\pi}{3} \right) $.  
        Within this interval, the solutions to $ \sin \theta = \frac{4}{5} $ are $ \theta = \arcsin \frac{4}{5} $ or $ \theta = \pi - \arcsin \frac{4}{5} $,
        but the latter is approximately $ \pi - 0.927 \approx 2.214 > \frac{2\pi}{3} \approx 2.094 $, which lies outside the interval; hence the only valid solution is $ \theta = \arcsin \frac{4}{5} \in \left( 0, \frac{\pi}{2} \right) $, and subsequently in this case, we have $\cos \theta = \frac{3}{5}$.  
        We then compute $ f(x_0 + 1) = 2\sqrt{3} \sin\left( \frac{\pi}{2} (x_0 + 1) + \frac{\pi}{3} \right) = 2\sqrt{3} \sin\left( \theta + \frac{\pi}{2} \right) = 2\sqrt{3} \cos \theta = 2\sqrt{3} \cdot \frac{3}{5} = \frac{6\sqrt{3}}{5} $.
        Therefore, the final result is: (1) $ \omega = \frac{\pi}{2} $; (2) $ f(x_0 + 1) = \frac{6\sqrt{3}}{5}$.
        \rule[-2ex]{0pt}{0pt} \\
        \hline
        \multicolumn{1}{c}{
        \raisebox{-0.28cm}{\includegraphics[width=0.75cm, height=0.75cm]{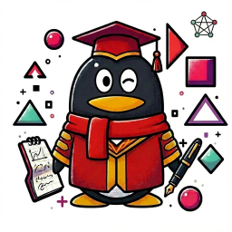}} \
        \rule{0pt}{3ex}\textbf{\ourapproach}
        \rule[-2ex]{0pt}{0pt}}\\
        \hline
        \rule{0pt}{3ex}
        In the Cartesian coordinate system, a sinusoidal curve is plotted, displaying a portion of the waveform within one period. The curve intersects the $x$-axis at point $B$ to the left of the $y$-axis and at point $C$ to the right of the $y$-axis, with the highest point of the curve marked as $A$. Points $A$, $B$, and $C$ are connected to form a triangle $\triangle ABC$, which the problem specifies is an equilateral triangle, and the side length of this triangle is equal to half the period of the trigonometric function.
        First, we need to simplify the analytical expression of the function $f(x)$ using trigonometric identities:
        $f(x) = 6 \cos^{2} \frac{\omega x}{2} + \sqrt{3} \sin \omega x - 3
        = \sqrt{3} \sin \omega x + 3 \cos \omega x
        = 2\sqrt{3} \sin \left( \omega x + \frac{\pi}{3} \right)$.
        From the simplified expression, the maximum value of the function is $2\sqrt{3}$, which corresponds to the ordinate of the highest point $A$, so $y_A = 2\sqrt{3}$. Since $\triangle ABC$ is an equilateral triangle and $BC$ lies on the $x$-axis, we have $2\sqrt{3} = \frac{\sqrt{3}}{2} BC \implies BC = 4$. Thus, the period of the trigonometric function is $T = 2 BC = 8$. Using the period formula $T = \frac{2\pi}{\omega}$, we get $\frac{2\pi}{\omega} = 8 \implies \omega = \frac{\pi}{4}$.
        From (1), we know the function is $f(x) = 2\sqrt{3} \sin \left( \frac{\pi}{4} x + \frac{\pi}{3} \right)$. Given $f(x_{0}) = \frac{8\sqrt{3}}{5}$, substituting this into the expression yields $\sin \left( \frac{\pi}{4} x_{0} + \frac{\pi}{3} \right) = \frac{4}{5}$.
        Next, we determine the range of the angle to find the sign of the cosine value. Given $x_{0} \in \left(\frac{-10}{3}, \frac{2}{3}\right)$, then $\alpha = \frac{\pi}{4} x_{0} + \frac{\pi}{3} \in \left(-\frac{\pi}{2}, \frac{\pi}{2}\right)$. Since $\sin \alpha = \frac{4}{5} > 0$ and $\alpha \in \left(-\frac{\pi}{2}, \frac{\pi}{2}\right)$, so the cosine value is $\cos \alpha = \frac{3}{5}$.
        Now we can calculate: $f(x_{0}+1) = 2\sqrt{3} \left[ \sin \left( \frac{\pi}{4} x_{0} + \frac{\pi}{3} \right) \cos \frac{\pi}{4} + \cos \left( \frac{\pi}{4} x_{0} + \frac{\pi}{3} \right) \sin \frac{\pi}{4} \right]$.
        Substituting the known values:
        $f(x_{0}+1) = 2\sqrt{3} \left( \frac{4}{5} \cdot \frac{\sqrt{2}}{2} + \frac{3}{5} \cdot \frac{\sqrt{2}}{2} \right) = \frac{7\sqrt{6}}{5}$.
        Final Answer:
        (1) $\omega = \frac{\pi}{4}$;
        (2) $f(x_{0}+1) = \frac{7\sqrt{6}}{5}$.
        \rule[-2ex]{0pt}{0pt} \\
        \hline
    \end{tabular}
    \caption{The elaborate response generated by prominent open‑source models.}
\end{table*}

\end{document}